\documentclass[%
 aip,
 amsmath,amssymb,
 reprint,%
]{revtex4-1}

\usepackage{graphicx}
\usepackage{dcolumn}
\usepackage{bm}
\usepackage{enumitem}
\usepackage[utf8]{inputenc}
\usepackage[T1]{fontenc}
\usepackage{mathptmx}
\usepackage{etoolbox}

\makeatletter
\def\@email#1#2{%
 \endgroup
 \patchcmd{\titleblock@produce}
  {\frontmatter@RRAPformat}
  {\frontmatter@RRAPformat{\produce@RRAP{*#1\href{mailto:#2}{#2}}}\frontmatter@RRAPformat}
  {}{}
}%
\makeatother
\usepackage[english]{babel}
\usepackage{microtype}
\usepackage{tabularx}
\newcolumntype{Y}{>{\centering\arraybackslash}X}

\usepackage{booktabs} 
\usepackage{multirow}
\usepackage{siunitx}

\usepackage{pgfplots}
\pgfplotsset{compat=newest}
\usepackage{tikz}
\usetikzlibrary{shapes.geometric, arrows.meta, positioning, fit, calc}

\usepackage{soul}

\usepackage{hyperref}

\hypersetup{
    colorlinks=true,
    citecolor=blue,
    linkcolor=blue,
    urlcolor=blue
}

\usepackage{caption}
\usepackage{subcaption}

\makeatletter
\def\section{\@startsection {section}{1}{\z@}{-2.0ex \@plus -0.5ex \@minus -0.2ex}{1.5ex \@plus 0.3ex}{\large\bfseries}}
\def\subsection{\@startsection{subsection}{2}{\z@}{-1.5ex \@plus -0.5ex \@minus -0.2ex}{1.0ex \@plus 0.2ex}{\normalsize\bfseries}}
\makeatother
\begin{document}

\preprint{AIP/123-QED}

\title[TripOptimizer]{TripOptimizer: Generative 3D Shape Optimization and Drag Prediction using Triplane VAE Networks}
\author{Parsa Vatani}
\email{parsa.vatani99@gmail.com}
\affiliation{Department of concepts and methods development in virtual fields, AUDI AG, Ingolstadt, Germany}
\affiliation{Technology and Bionics Faculty, Rhine-Waal University of Applied Sciences, Kleve, Germany}

\author{Mohamed Elrefaie}
\affiliation{Department of Mechanical Engineering, Massachusetts Institute of Technology, Cambridge, MA, USA}
\affiliation{Schwarzman College of Computing, Massachusetts Institute of Technology, Cambridge, MA, USA}

\author{Farhad Nazarpour}
\affiliation{Department of concepts and methods development in virtual fields, AUDI AG, Ingolstadt, Germany}

\author{Faez Ahmed}
\affiliation{Department of Mechanical Engineering, Massachusetts Institute of Technology, Cambridge, MA, USA}

\date{\today}

\begin{abstract}
The computational cost of traditional Computational Fluid Dynamics-based Aerodynamic Shape Optimization severely restricts design space exploration. This paper introduces TripOptimizer, a fully differentiable deep learning framework for rapid aerodynamic analysis and shape optimization directly from vehicle point cloud data. TripOptimizer employs a Variational Autoencoder featuring a triplane-based implicit neural representation for high-fidelity 3D geometry reconstruction and a drag coefficient prediction head. Trained on DrivAerNet++, a large-scale dataset of 8,000 unique vehicle geometries with corresponding drag coefficients computed via Reynolds-Averaged Navier-Stokes simulations, the model learns a latent representation that encodes aerodynamically salient geometric features. We propose an optimization strategy that modifies a subset of the encoder parameters to steer an initial geometry towards a target drag value, and demonstrate its efficacy in case studies where optimized designs achieved drag coefficient reductions up to 11.8\%. These results were subsequently validated by using independent, high-fidelity Computational Fluid Dynamics simulations with more than 150 million cells. A key advantage of the implicit representation is its inherent robustness to geometric imperfections, enabling optimization of non-watertight meshes, a significant challenge for traditional adjoint-based methods. The framework enables a more agile Aerodynamic Shape Optimization workflow, reducing reliance on computationally intensive CFD simulations, especially during early design stages.

\end{abstract}

\maketitle

\section{Introduction}
Vehicle aerodynamics critically determines energy efficiency and performance, with aerodynamic drag being a dominant resistive force at highway speeds \cite{Hucho1998, Katz2001}. Conventional Aerodynamic Shape Optimization (ASO) relies on iterative design modification and computationally intensive Computational Fluid Dynamics (CFD) simulations. The substantial cost and time of these methods constrain design space exploration, particularly in early vehicle development \cite{smith2006evaluation}. Data-driven deep learning approaches have emerged as powerful alternatives.

\begin{figure*}[!htbp]
    \centering
    \includegraphics[width=\textwidth]{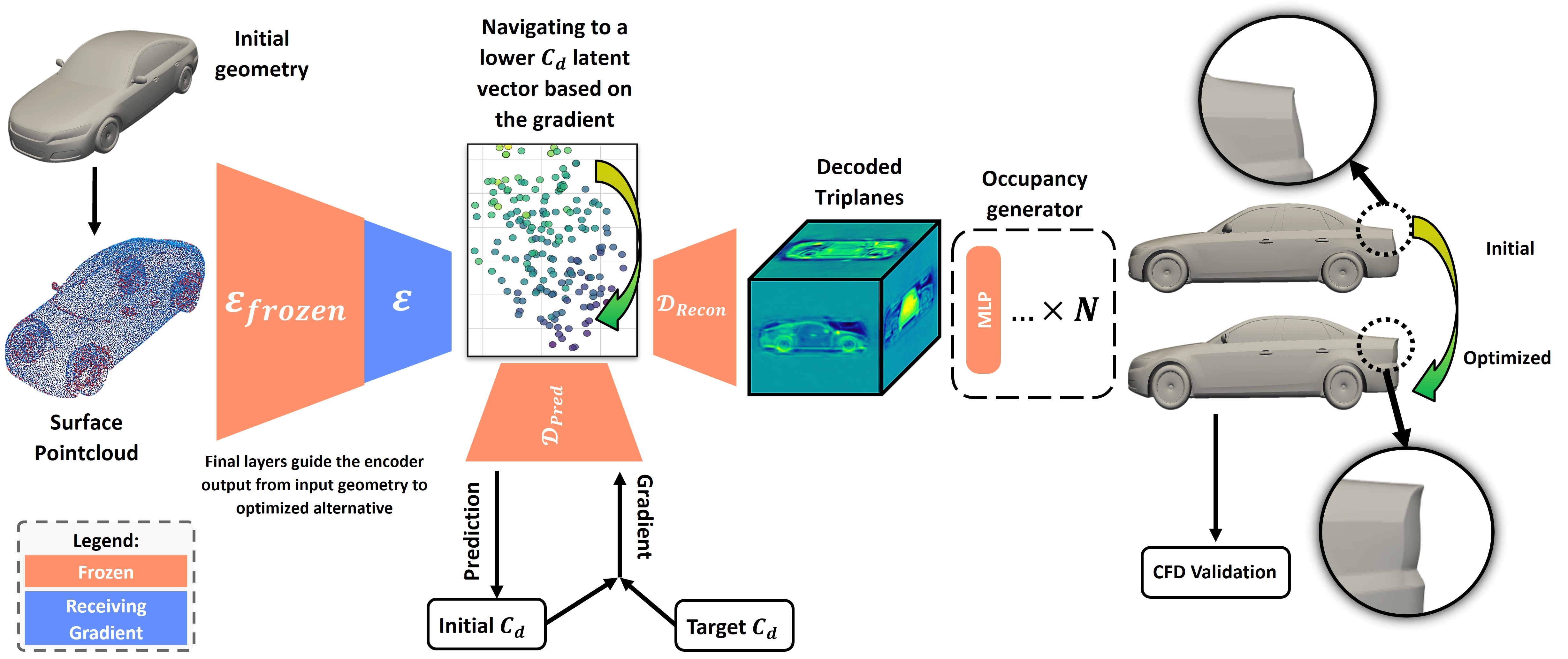}
\caption{Overview of our proposed TripOptimizer, a fully differentiable generative triplane-based model for aerodynamic analysis and shape optimization. The model jointly reconstructs high-fidelity 3D vehicle shapes and predicts their drag coefficients ($C_d$), then optimizes designs towards user-defined aerodynamic targets.}
    \label{fig:framework_overview}
\end{figure*}

This paper introduces a novel deep learning framework for accelerated aerodynamic analysis and performance-driven shape modification operating directly on vehicle point clouds. The core of this work is a unified Variational Autoencoder (VAE)~\cite{Kingma2013VAE} architecture that performs two concurrent tasks: the high-fidelity reconstruction of the 3D shape and the accurate prediction of the aerodynamic drag coefficient ($C_d$). For geometry generation, the model leverages a triplane-based implicit neural representation to produce continuous and detailed surfaces \cite{Chan2022EG3D, Peng2020ConvolutionalOccupancy}. The framework's primary novelty lies in its shape optimization methodology. Instead of directly manipulating a design's latent code, the proposed strategy modifies a subset of the VAE's encoder parameters to efficiently guide an initial vehicle geometry towards a user-defined target $C_d$. A key contribution is the framework's ability to handle imperfect geometries; its point cloud input and implicit representation make it robust to non-watertight meshes, a notable limitation of traditional adjoint-based CFD optimization. Significant drag reductions were achieved for the optimized shapes and subsequently validated through independent, high-fidelity CFD simulations utilizing meshes of up to 150 million cells. The complete framework is trained and validated on the DrivAerNet++ dataset \cite{Elrefaie2024DrivAerNetPP, elrefaie2024drivaernet, elrefaie2025drivaernet}, a large-scale collection of diverse vehicle morphologies with their corresponding CFD-computed $C_d$ values.

The presented approach can be implemented in the initial automotive development phase for aerodynamic assessment and design improvement, shortening engineering time and lowering development costs. The method is particularly advantageous in early design stages where quick $C_d$ estimation and optimization guidance can eliminate hours of simulation, leading to a high-performance final candidate. For instance, a 10\% reduction in aerodynamic drag can increase an electric vehicle's range by approximately 5\%. For internal combustion engine vehicles, a similar 10\% drag reduction can improve fuel efficiency by approximately 3-5\% during highway driving \cite{schuetz2015aerodynamics}.

In order to explain our approach and results, this paper is organized as follows: Section~\ref{sec:related_work} reviews prior work in the field. Section~\ref{sec:methodology} details our proposed methodology. Section~\ref{sec:results_discussion_revised} presents and discusses the experimental results, followed by the conclusion and future work in Sections~\ref{sec:conclusion} and \ref{sec:future_work}, respectively.

\section{Related Works}
\label{sec:related_work}
The pursuit of aerodynamically performant vehicle designs has traditionally been an iterative and computationally intensive process. This section reviews prior work in three key areas that form the foundation for present work's proposed framework: aerodynamic shape optimization and surrogate models, 3D geometric representations for deep learning, and generative models for design exploration.

\subsection{Aerodynamic Shape Optimization and Surrogate Models}
Historically, Aerodynamic shape optimization has relied on coupling Computational Fluid Dynamics solvers with numerical optimization algorithms \cite{Martins2021}. Gradient-based methods, which often employ adjoint sensitivities, can efficiently handle a large number of design variables but necessitate multiple, resource-intensive CFD evaluations \cite{lavimi2024review}. The prohibitive computational cost of direct CFD-driven optimization motivated the development of surrogate models, also known as meta-models. These surrogate models approximate the input-output relationship of the expensive simulations. Early and widely adopted surrogate techniques included statistical methods like Kriging (Gaussian Process Regression) and polynomial regression \cite{Forrester2008}. More recently, deep neural networks (DNNs) have gained prominence due to their capacity for modeling the highly non-linear relationships inherent in fluid dynamics \cite{Zhang2021Review}. Researchers have successfully applied DNNs to predict scalar aerodynamic coefficients, such as drag and lift, and to reconstruct entire flow fields \cite{Bhatnagar2019Prediction, Thuerey2020}. To further improve data efficiency, multi-fidelity modeling approaches have been developed. The multi-fidelity methods combine information from simulations of varying accuracy in order to build robust surrogates with a limited budget of high-fidelity CFD runs \cite{Peherstorfer2018, Han2012Surrogate}.
\subsection{Three-Dimensional Geometric Representations for Deep Learning}
The choice of 3D geometric representation is an important step when applying deep learning to ASO. The importance arises from the varying information that each representation holds. Point clouds, defined as unstructured sets of 3D coordinates, offer a direct and flexible representation of complex surfaces while being computationally inexpensive \cite{Guo2020DeepLearning3DReview}. Pioneering work by Qi et al. \cite{Qi2017PointNet, Qi2017PointNet++} established learning methods for applying deep learning directly to point sets. For aerodynamic applications, autoencoders operating on point clouds have been explored for learning shape representations suitable for performance prediction \cite{Achlioptas2018LearningPointClouds}. Mesh-based representations have gained interest because of the connectivity information they hold. Graph Neural Networks (GNNs) are one of the methods fit for such data. GNN models such as MeshGraphNets have demonstrated the ability to learn physics simulations directly on meshes \cite{Pfaff2020}. However, the computational and memory costs associated with processing the large graphs (typically used in industrial CFD meshes) remain a significant bottleneck for scalability. Voxel grids offer an alternative by discretizing the 3D space, which allows for the use of standard 3D Convolutional Neural Networks (CNNs) \cite{Wu20163DGAN}. Their principal drawback is that the memory consumption scales cubically with the voxel resolution. This drawback makes the high resolutions required for fine geometric details computationally prohibitive. Implicit Neural Representations (INRs) define shapes as continuous functions that map coordinates to properties such as signed distance or occupancy \cite{Park2019DeepSDF, Mescheder2019Occupancy}. This approach is memory-efficient, and it can represent shapes at arbitrary fidelity. The triplane representation is a notable hybrid INR. Triplane methods factorize a 3D field into three orthogonal 2D feature planes, which are then queried by a small decoding MLP \cite{Chan2022EG3D, Peng2020ConvolutionalOccupancy, Gao2022GET3D}. Triplanes have been successfully used for high-quality generative tasks \cite{Eric2021TriVol, Shue20233DStyleDiffusion, Saha2024TriplaneGaussian}. The TripNet~\cite{tripnet} method also utilized triplanes for aerodynamic prediction through a multi-stage process. In this approach, triplanes are first generated and pre-computed by fitting them to the occupancy field of a given vehicle geometry. These static, pre-computed triplane representations are then used as a fixed input for a separate regression network tasked with predicting the aerodynamic coefficients.


\subsection{Generative Models and Latent Space Optimization}
Generative models, such as VAEs \cite{Kingma2013VAE} and Generative Adversarial Networks (GANs) \cite{Wu20163DGAN}, learn to synthesize novel data by mapping from a low-dimensional latent space. This learned latent space often captures meaningful semantic variations of the training data, providing a powerful tool for design exploration and optimization \cite{GomezBombarelli2018, SanchezLengeling2018, Oh2019DeepGenStruct}. Optimization in this context involves searching the latent space for vectors $z$ that correspond to designs with desired properties. This search can be guided by various techniques, including gradient-based methods, Bayesian optimization, or evolutionary algorithms \cite{Bao2021, liu2023deep}. A central challenge is to ensure the latent space is well-structured, which means smooth and continuously interpolable. Another important consideration is that its dimensions correlate with meaningful design attributes. This is often addressed through VAE regularization techniques, such as $\beta$-VAE \cite{Higgins2017betaVAE}, or by promoting disentanglement \cite{Mathieu2019Disentangling, Chen2018IsolatingFactors}. Recent work by Tran et al. employed a VAE-based generative model for aerodynamic optimization, using a voxel-based representation of the vehicle geometry and an MLP-based autoencoder \cite{Tran2024}. This methodology presents key limitations. The voxel-based approach can result in the loss of critical, high-frequency geometric features. Furthermore, MLP-based VAEs may not adequately preserve the complex, high-dimensional information inherent in 3D vehicle shapes. TripOptimizer addresses these shortcomings, namely the loss of geometric quality in voxel-based and MLP approaches, by utilizing a 3D representation that preserves fine geometric details and enables a more effective generative model for aerodynamic analysis.

\section{Methodology}
\label{sec:methodology}
The methodology in this study introduces \textbf{TripOptimizer}, a framework for aerodynamic shape optimization built upon a Variational Autoencoder (VAE). The VAE is trained to process vehicle geometries and learn a compact latent representation from point cloud data. Point clouds are selected as the input modality for their raw format and simple extraction from CAD models. This learned representation facilitates both 3D shape reconstruction and drag coefficient prediction. A core distinction of the present work lies in its internal geometric representation. The VAE's decoder generates triplanes from a latent code $z$, which the encoder produces from the input point cloud. This design enables end-to-end learning within a unified latent space. An overview of the complete framework is depicted in Figure~\ref{fig:framework_overview}.
This learned manifold is the basis for our novel optimization strategy. Instead of directly manipulating a shape's latent vector $z$, our method fine-tunes a subset of the encoder's parameters. This adjustment alters how the encoder maps a fixed initial shape $X_{init}$ to a new latent representation $z_{opt}$. The new representation is optimized to correspond to a desired target $C_d$. This process constitutes a targeted adaptation of the learned geometric prior, guided by the aerodynamic objective. The approach leverages the VAE's understanding of valid vehicle geometries to generate coherent and plausible shape modifications that meet the performance criteria.

\subsection{Dataset Overview}
The deep learning framework presented in this study was trained and validated utilizing the DrivAerNet++ dataset, a comprehensive and large-scale resource developed specifically for data-driven aerodynamic research \cite{Elrefaie2024DrivAerNetPP}. This dataset provides an ideal foundation for training robust generative and predictive models, owing to its size, diversity, and fidelity. It contains 8,000 unique vehicle designs, each with corresponding aerodynamic coefficients derived from validated CFD simulations. The geometries are procedurally generated by morphing parametric models based on the industry-standard DrivAer reference vehicle. This method ensures that all designs are physically plausible and relevant to real-world automotive engineering.

\begin{figure}
   \centering    \includegraphics[width=\columnwidth]{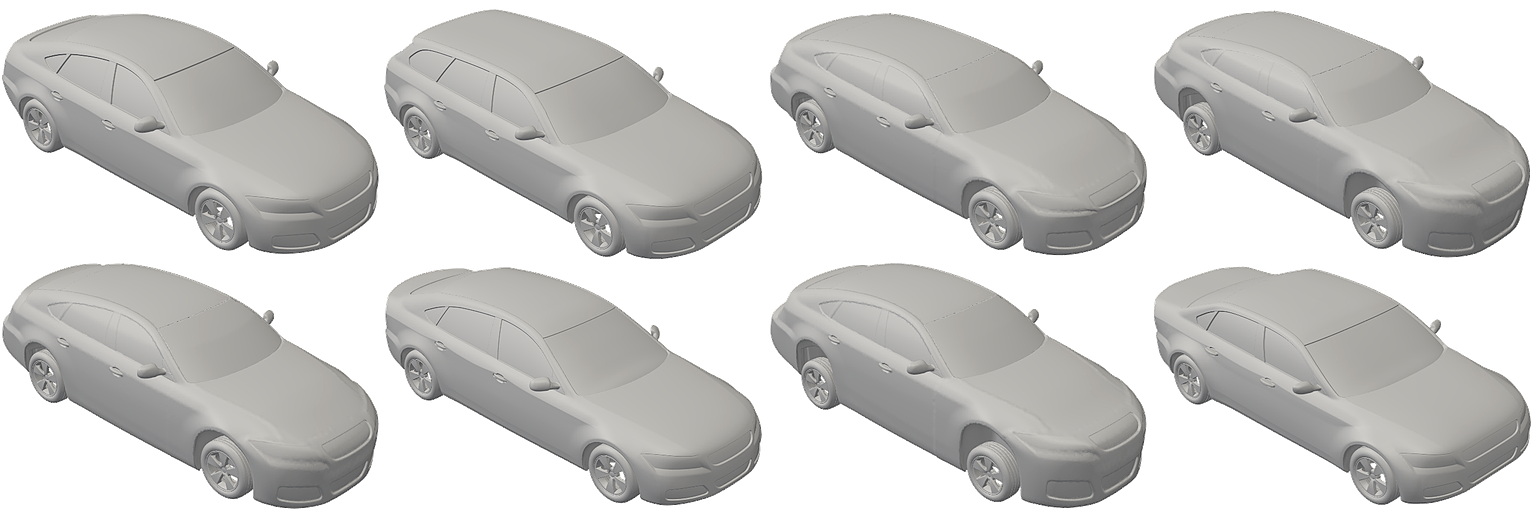}
\caption{A selection of diverse vehicle shapes within the DrivAerNet++ dataset~\cite{Elrefaie2024DrivAerNetPP}. The figure covers Estatebacks, Notchbacks, and Fastbacks.}
    \label{fig:dataset_samples}
\end{figure}

A principal strength of DrivAerNet++ is its extensive geometric diversity, which is essential for training models capable of generalizing across a wide design space. The dataset systematically covers three primary vehicle typologies: fastback, notchback, and estateback. To further enhance its applicability to modern vehicle development, the dataset also incorporates crucial design variations. These include detailed underbodies, which are characteristic of internal combustion engine vehicles (ICE), and smooth underbodies, which are common in electric vehicles (EV). Furthermore, a variety of wheel designs are represented, spanning open, closed, detailed, and smooth styles.  The span of this design space is further illustrated by the sample of geometries shown in Figure \ref{fig:dataset_samples}. This geometric variability results in a broad and complex distribution of aerodynamic performance.

\subsection{Data Conversion and Preprocessing}
\label{sec:data_preprocessing}
The data preprocessing pipeline converts each raw STL mesh into a fixed-size point cloud, which serves as the input for our TripOptimizer. To ensure consistent input dimensionality, a total of $N$ points are sampled from each vehicle's surface using a hybrid strategy that preserves critical geometric details. First, Farthest Point Sampling (FPS)\cite{Qi2017PointNet++} is employed to select 75\% of the points, guaranteeing a uniform and broad representation of the vehicle's surface. The remaining 25\% are specifically sampled from regions exhibiting high curvature and along sharp edges. This focused sampling ensures that aerodynamically significant features are captured with higher fidelity. For the final training configuration, the point cloud size $N$ is set to 50,000. A comprehensive list of data preprocessing hyperparameters is provided in Appendix \ref{app:hyperparams} (Table \ref{tab:data_preprocessing_hyperparams}). A more granular description of the entire preprocessing pipeline, including mesh decimation and sampling strategies, is provided in Appendix \ref{app:preprocessing}.

\begin{figure}
    \centering
    \includegraphics[width=\linewidth]{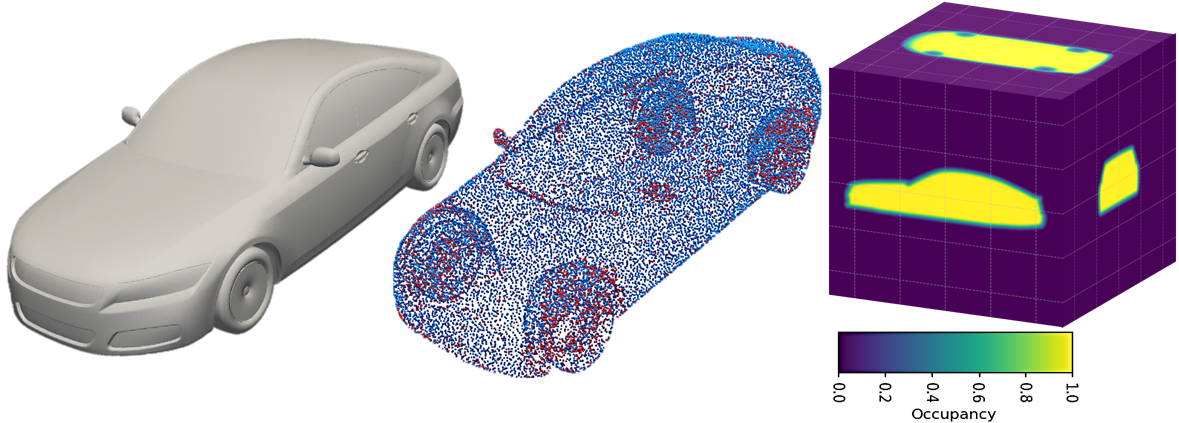}
    \caption{The data transformation visualization, with the original geometry (left), corresponding surface point cloud (middle), and three orthogonal slices of semi-continuous occupancy field (right).}
    \label{fig:data_representation_stages}
\end{figure}

In order to generate the ground truth data required for supervising the VAE's geometric reconstruction, a dense, semi-continuous occupancy field is computed for each vehicle. The procedure commences with a normalization step, where each geometry is centered at the origin and its bounding box is scaled to a unit length. Following normalization, the Signed Distance Function (SDF) value is calculated for a large set of $M_{total}$ query points (e.g., 1.28 million, see Table \ref{tab:data_preprocessing_hyperparams}). These points are sampled using a mixed strategy to ensure comprehensive spatial coverage. A significant portion of the points is sampled uniformly within the bounding box in order to have uniform occupancy information. For the purpose of capturing fine-grained details, additional points are sampled both directly on the mesh surface and in its immediate vicinity. The latter is achieved by perturbing surface points along their normals with Gaussian noise. For each query point, the unsigned distance to the nearest geometry surface is efficiently computed using a KDTree\cite{bentley1975multidimensional} built from the mesh vertices. The sign, which indicates whether a point is inside or outside the mesh, is robustly determined via a ray-casting method that performs a majority vote based on the parity of ray-to-mesh intersections. These SDF values are transformed into semi-continuous ground truth occupancy values, $o_{true,i} \in [0,1]$, using the following function:
\begin{equation} \label{eq:sdf_to_occ_revised_main}
o_{true,i}(x_{q,i}) = \text{clip}_{[0,1]}\left(0.5 - 0.5 \cdot \frac{\text{SDF}(x_{q,i})}{s}\right),
\end{equation}
where $o_{true,i}$ is the occupancy for query point $x_{q,i}$. The parameter $s$ is a small positive threshold proportional to the mesh's characteristic size, which defines the width of the smooth transition region near the surface. The complete set of $M_{total}$ query points and their occupancy values is stored for each vehicle. For each vehicle, the corresponding $C_d$ value, sourced from a metadata file, is also stored.  During VAE training, this large set is dynamically subsampled to $M$ points per shape in each iteration for computational efficiency.
An example of these data transformations is shown in Figure \ref{fig:data_representation_stages}.

\subsection{Model Architecture}
The VAE architecture (shown in Figure \ref{fig:framework_overview}) consists of three main parts: a point cloud encoder, a latent space, and two decoding heads; one for 3D shape reconstruction via triplanes and another for $C_d$ prediction. The specific architectural hyperparameters for each component are detailed in Appendix \ref{app:hyperparams} (Table \ref{tab:model_architecture_hyperparams}).

\begin{figure}[!htbp]
    \centering
    \includegraphics[width=\linewidth]{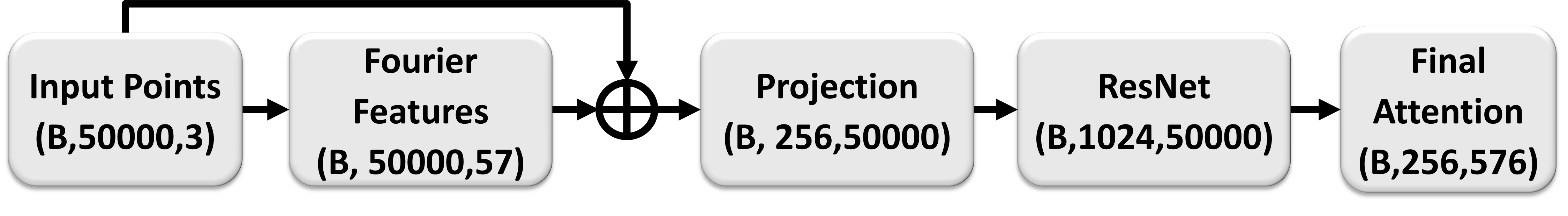}
    \caption{Data transformation within the Encoder alongside output dimensions for 50k points input. }
    \label{fig:encoder_dataflow}
\end{figure}

\textbf{Point Cloud Encoder:}
The encoder network maps an input surface point cloud $X \in \mathbb{R}^{N \times 3}$ to the parameters of a diagonal Gaussian distribution in the latent space. Each 3D point coordinate $p \in \mathbb{R}^3$ is augmented using Fourier positional encodings \cite{Mildenhall2020NeRF}, $\gamma(p)$, to capture high-frequency details:
\begin{multline}
\label{eq:fourier_features_model_main}
\gamma(p) = (p, \sin(2^0 \pi p), \cos(2^0 \pi p), \dots, \\ \sin(2^{L-1} \pi p), \cos(2^{L-1} \pi p)),
\end{multline}
where $L$ is the number of frequency bands. The augmented features are processed by residual blocks employing 1D convolutions, instance normalization, and ReLU activations \cite{He2016DeepResidual}. An attention-based module\cite{hu2021a2} then aggregates these point features using a grid of learnable query vectors ($H_{latent} \times W_{latent}$) to produce parameters (mean $\mu_\phi(X)$ and log-variance $\log \sigma^2_\phi(X)$) for the approximate posterior $q_\phi(z|X) = \mathcal{N}(z; \mu_\phi(X), \text{diag}(\sigma^2_\phi(X)))$. The latent variable $z \in \mathbb{R}^{D_{latent} \times H_{latent} \times W_{latent}}$ is sampled using reparameterization:
\begin{equation}
\label{eq:reparam_trick_model_main}
z = \mu_\phi(X) + \sigma_\phi(X) \odot \epsilon, \quad \text{where } \epsilon \sim \mathcal{N}(0, I),
\end{equation}
and $\sigma_\phi(X) = \exp(0.5 \log \sigma^2_\phi(X))$ is the standard deviation. Figure~\ref{fig:encoder_dataflow} illustrates the encoder dataflow and output shapes.

\begin{figure}[!htbp]
    \centering
    \includegraphics[width=\linewidth]{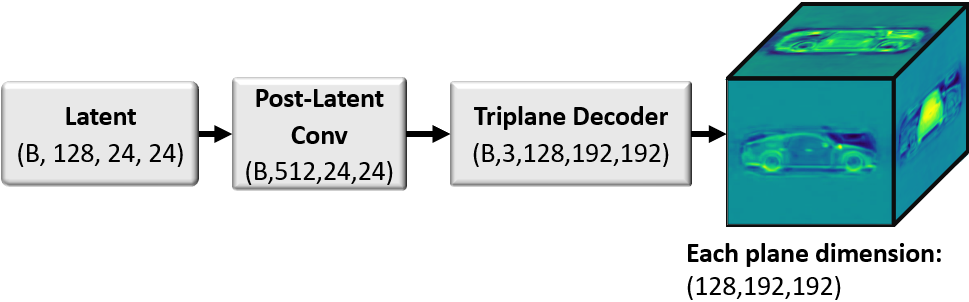}
    \caption{Visualization of the Triplane-based Occupancy Decoder components and example triplane of geometry ID F-S-WWC-WM-260 from the DrivAerNet++ dataset.}
    \label{fig:triplane_decoder_viz}
\end{figure}

\textbf{Occupancy Decoder via Triplane Representation:}
The shape decoder reconstructs the 3D geometry from the latent sample $z$. The latent grid $z$ is projected by a 1x1 convolution and fed to a network inspired by U-Net \cite{Ronneberger2015Unet}, in order to generate the triplanes. This network produces three orthogonal 2D feature planes $F_{xy}, F_{xz}, F_{yz}$, each with $C_{plane}$ feature channels. To predict occupancy $o_{pred,k}$ for a query point $x_{q,k}=(x_k,y_k,z_k)$, its coordinates are normalized and scaled by a box warp factor $\alpha_{box}$, which is set to a value of 1.1. Features $f_{xy,k}, f_{xz,k}, f_{yz,k}$ are bilinearly sampled from the respective planes. These are concatenated to $f_{cat,k}$ and processed by an MLP with a final sigmoid activation:
\begin{equation}
o_{pred,k} = \text{sigmoid}(\text{MLP}_{occ}(f_{cat,k}; \theta_{occ})),
\end{equation}
where $\theta_{occ}$ are the MLP parameters. This defines a continuous occupancy field \cite{Chan2022EG3D, Gao2022GET3D}. Figure~\ref{fig:triplane_decoder_viz} visualizes the triplane decoder and a sample set of feature planes. After obtaining the semi-continuous occupancy field, the iso-surface of the geometry will be extracted using marching cubes \cite{lorensen1998marching}  and a smoothing step.

\textbf{$C_d$ Prediction Head:}
This head predicts $C_d$ from the latent grid $z$. It uses 2D convolutional layers, with Squeeze-and-Excitation (SE) blocks \cite{Hu2018SqueezeExcitation}, to extract spatial features. These are flattened and processed by a multi-head self-attention mechanism\cite{vaswani2017attention}:
\begin{equation}
\label{eq:attention_cd_head}
\text{Attention}(Q, K, V) = \text{softmax}\left(\frac{QK^T}{\sqrt{d_k}}\right)V,
\end{equation}
where $Q, K, V$ are query, key, and value projections of the input sequence, and $d_k$ is the key dimension. The attention output is aggregated (e.g., mean-pooled) and passed through an MLP to regress $C_{d,pred}$.

\begin{figure*}
    \centering
    \includegraphics[width=\textwidth]{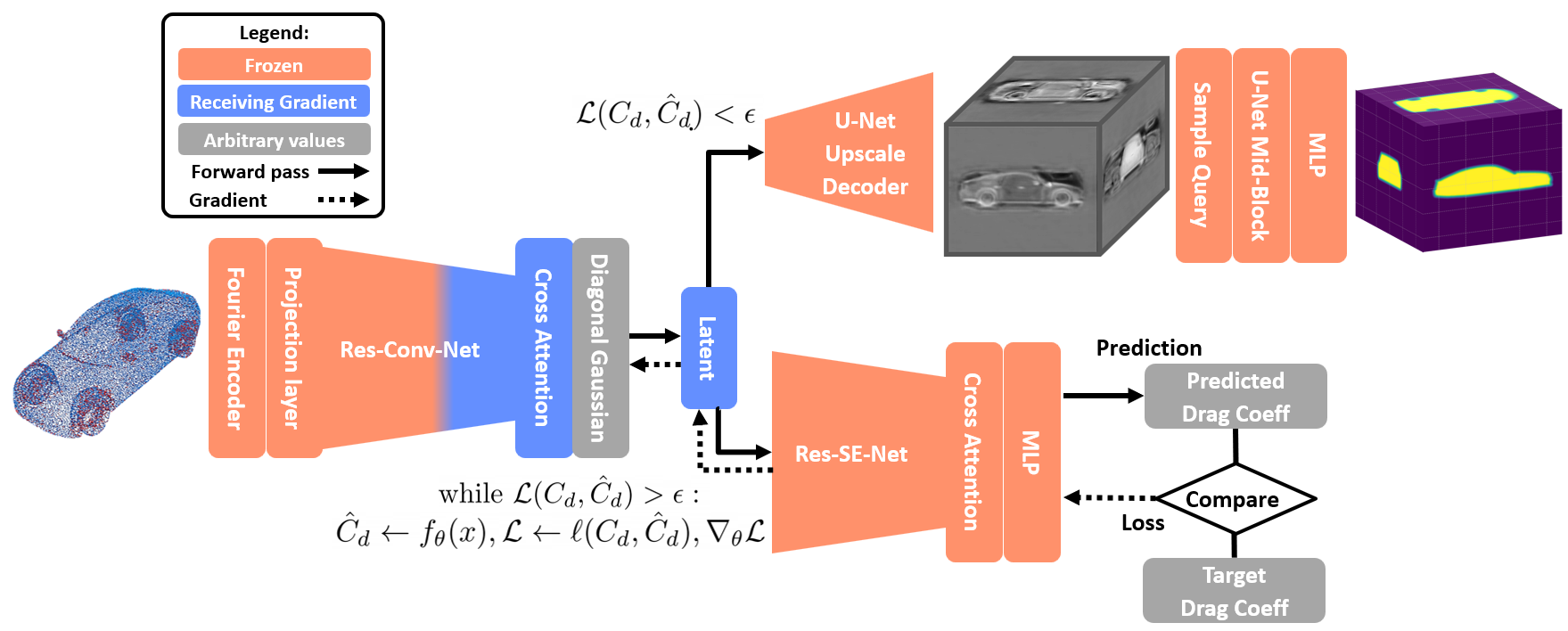}
    \caption{Diagram of the aerodynamic shape optimization process through encoder parameter modification.}
    \label{fig:optimization_process}
\end{figure*}

\subsection{Training Strategy}
The VAE model is trained end-to-end by minimizing a composite loss function $L_{total}$. Its performance is evaluated using standard metrics for drag coefficient prediction, including the coefficient of determination ($R^2$), Mean Absolute Error (MAE), Root Mean Squared Error (RMSE), and Mean Absolute Percentage Error (MAPE), as well as the F1 score for geometric reconstruction fidelity, with the formulas for all metrics provided in Appendix \ref{app:metrics}. The training data is partitioned into training, validation, and test sets according to predefined splits loaded from a JSON file. The specific hyperparameters governing the training process are detailed in Appendix \ref{app:hyperparams} (Table \ref{tab:training_hyperparams}). The total loss $L_{total}$ is a weighted sum of several components, each designed to guide the learning process towards specific objectives:

\textbf{Reconstruction Loss ($L_{recon}$)}: This loss ensures that the decoder can accurately reconstruct the input vehicle's geometry from its latent representation. It measures the discrepancy between the predicted occupancy values $o_{pred,i}$ and the ground truth occupancy values $o_{true,i}$ for a batch of $M$ query points. The Smooth L1 loss\cite{girshick2015fast} is employed, as it is less sensitive to outliers than Mean Squared Error (MSE) while providing smooth gradients for small errors. It is defined as:
\begin{equation}
L_{recon} = \frac{1}{M} \sum_{i=1}^M \mathcal{L}_{smoothL1}(o_{pred,i}, o_{true,i}; \beta_{occ}),
\end{equation}
where $\beta_{occ}$ is a hyperparameter controlling the transition point.

\textbf{Drag Coefficient Prediction Loss ($L_{C_d}$)}: This term penalizes deviations between the predicted drag coefficient $C_{d,pred}$ and the ground truth drag coefficient $C_{d,true}$. Similar to the reconstruction loss, the Smooth L1 loss is used for its robustness:
\begin{equation}
L_{C_d} = \mathcal{L}_{smoothL1}(C_{d,pred}, C_{d,true}; \beta_{cd}),
\end{equation}

\textbf{Kullback-Leibler (KL) Divergence Loss ($L_{KL}$)}: This is a crucial component of VAEs, acting as a regularizer on the latent space. It encourages the learned approximate posterior distribution $q_\phi(z|X)$ (parameterized by the encoder) to be close to a chosen prior distribution $p(z)$, typically a standard multivariate Gaussian $\mathcal{N}(0,I)$. For diagonal Gaussian distributions, where the encoder outputs mean $\mu_{\phi,j}(X)$ and log-variance $\log \sigma^2_{\phi,j}(X)$ for each latent dimension $j$, the KL divergence is calculated analytically as:
\begin{equation}
\begin{split}
L_{KL} &= D_{KL}(q_\phi(z|X) || p(z)) \\
&= \frac{1}{2} \sum_{j=1}^{D_{latent}'} \left( \sigma^2_{\phi,j}(X) + \mu^2_{\phi,j}(X) - 1 - \log \sigma^2_{\phi,j}(X) \right),
\end{split}
\end{equation}
where the sum is over all dimensions of the flattened latent space $D_{latent}' = D_{latent} \cdot H_{latent} \cdot W_{latent}$. This loss promotes a smooth and continuous latent space. The weight $w_{KL}$ applied to this term is annealed during training (e.g., linearly increasing over the first 5 epochs).

\textbf{Boundary Loss ($L_{boundary}$)}: To encourage sharper surface definitions, an optional boundary loss can be included. This loss targets query points $x_{q,i}$ that lie close to the true surface of the vehicle (where $o_{true,i}$ is near 0.5). For these points, the loss penalizes predictions $o_{pred,i}$ that are not also close to 0.5:
\begin{equation}
L_{boundary} = \frac{1}{M_{boundary}} \sum_{i \in S_{boundary}} (o_{pred,i} - 0.5)^2,
\end{equation}
where $S_{boundary}$ is the set of query points near the surface and $M_{boundary}$ is its count.

\textbf{Triplane L1 Regularization ($L_{triplane\_L1}$)}: To encourage sparsity or smoothness in the generated triplane features $F_{xy}, F_{xz}, F_{yz}$, an L1 regularization term can be applied:
\begin{equation}
L_{triplane\_L1} = \frac{1}{N_{elements}} \sum_{P \in \{F_{xy}, F_{xz}, F_{yz}\}} \sum_{u,v,c} |P_{u,v,c}|,
\end{equation}
where $|P_{u,v,c}|$ is the absolute value of a feature element in plane $P$, and $N_{elements}$ is the total number of feature elements across all three planes.

The \textbf{Total Loss Function} is a weighted sum of these components:
\begin{multline}
L_{total} = w_{recon}L_{recon} + w_{C_d}L_{C_d} + w_{KL}L_{KL} \\
+ w_{boundary}L_{boundary} + w_{triplane\_L1}L_{triplane\_L1}.
\end{multline}
The weights balance the influence of each objective. The model is trained using the AdamW optimizer \cite{Loshchilov2017AdamW}.
\subsection{Aerodynamic Shape Optimization via Encoder Parameter Refinement}
Once the VAE is trained, its learned structure is exploited for aerodynamic shape optimization. A common approach for such tasks is to perform gradient-based optimization directly on the latent vector $z$. However, such a traversal may navigate into sparsely populated or poorly defined regions of the latent space, potentially yielding decoded geometries that lack coherence or contain artifacts. To mitigate this, we propose an alternative strategy that modifies the mapping from the input shape to the latent space, rather than the latent code itself. Given an initial vehicle design $X_{init}$ (as a point cloud), the objective is to modify its shape to achieve a user-specified target drag coefficient $C_d^{target}$.
The optimization strategy, outlined in Figure \ref{fig:optimization_process}, involves fine-tuning parts of the encoder network. This approach is posited to better preserve the design's core identity by re-interpreting the fixed input geometry through a modified perceptual lens, thereby constraining the search to plausible variations within the learned manifold. 
The process begins by encoding the initial design $X_{init}$ using the trained encoder $\phi$ to obtain its latent mode $z_{init} = \mu_\phi(X_{init})$. The corresponding initial predicted drag is $C_{d}^{init}$, computed by the $C_d$ prediction head (with parameters $\theta_{cd}$) from $z_{init}$.
During optimization, the parameters of the occupancy decoder ($\theta_{occ}$) and the $C_d$ prediction head ($\theta_{cd}$) are kept frozen.
A copy of the original encoder parameters, denoted $\phi'$, is created. A subset of these parameters is made trainable. Specifically, a user-defined number of the initial layers of this encoder $\phi'$ are kept frozen, while the subsequent, typically deeper, layers are allowed to be updated.
The optimization objective is to find a set of modified trainable encoder parameters $\phi'_{final}$ that minimizes the squared difference between the $C_d$ predicted for the fixed input $X_{init}$ (when encoded by the modified encoder $\phi'$) and the target $C_d^{target}$:
\begin{equation}
L_{opt}(\phi') = \left( \text{Head}_{C_d}(\mu_{\phi'}(X_{init}); \theta_{cd}) - C_d^{target} \right)^2.
\end{equation}
The input point cloud $X_{init}$ remains unchanged; only how $\phi'$ interprets this input is altered. The gradients $\nabla_{\phi'} L_{opt}(\phi')$ are computed for the trainable parameters in $\phi'$.
An optimizer, such as AdamW (with learning rate $\eta$), iteratively updates these parameters for a predefined number of steps:
\begin{equation}
\phi'_{new} = \phi'_{old} - \eta \nabla_{\phi'} L_{opt}(\phi'_{old}).
\end{equation}
The extent of geometric modification is controlled by the number of initial encoder layers that are frozen.
Once converged, the final $\phi'_{final}$ re-encodes $X_{init}$ to an optimized latent representation $z_{opt} = \mu_{\phi'_{final}}(X_{init})$. This $z_{opt}$ is decoded to the optimized 3D shape using the frozen occupancy decoder, typically by extracting an iso-surface. This process was applied to specific vehicle geometries to demonstrate its efficacy, as detailed in Section \ref{sec:optimization_case_study}.

\section{Results and Discussion}
\label{sec:results_discussion_revised}

The VAE framework's performance was assessed on its dual capabilities: predicting the aerodynamic $C_d$ and reconstructing 3D vehicle geometry. Industry-scale CFD simulations with more than 150M meshes were conducted to validate the optimization results. Variations in training methodology and data representations were also examined to determine their influence on these outcomes. The quantitative metrics are presented in Tables \ref{tab:cd_prediction_results_main_revised} and \ref{tab:training_ablation_results_main_revised}, while qualitative insights are drawn from analyses of error distributions, reconstruction quality, latent space organization, and an optimization case study, informed by visualizations such as those depicted in Figures \ref{fig:performance_combined} through \ref{fig:optimization_signed_distance}.

\subsection{Drag Coefficient Prediction Accuracy}

\begin{figure*}
    \centering
    \begin{subfigure}[t]{0.666\linewidth}
        \centering
        \includegraphics[width=\linewidth]{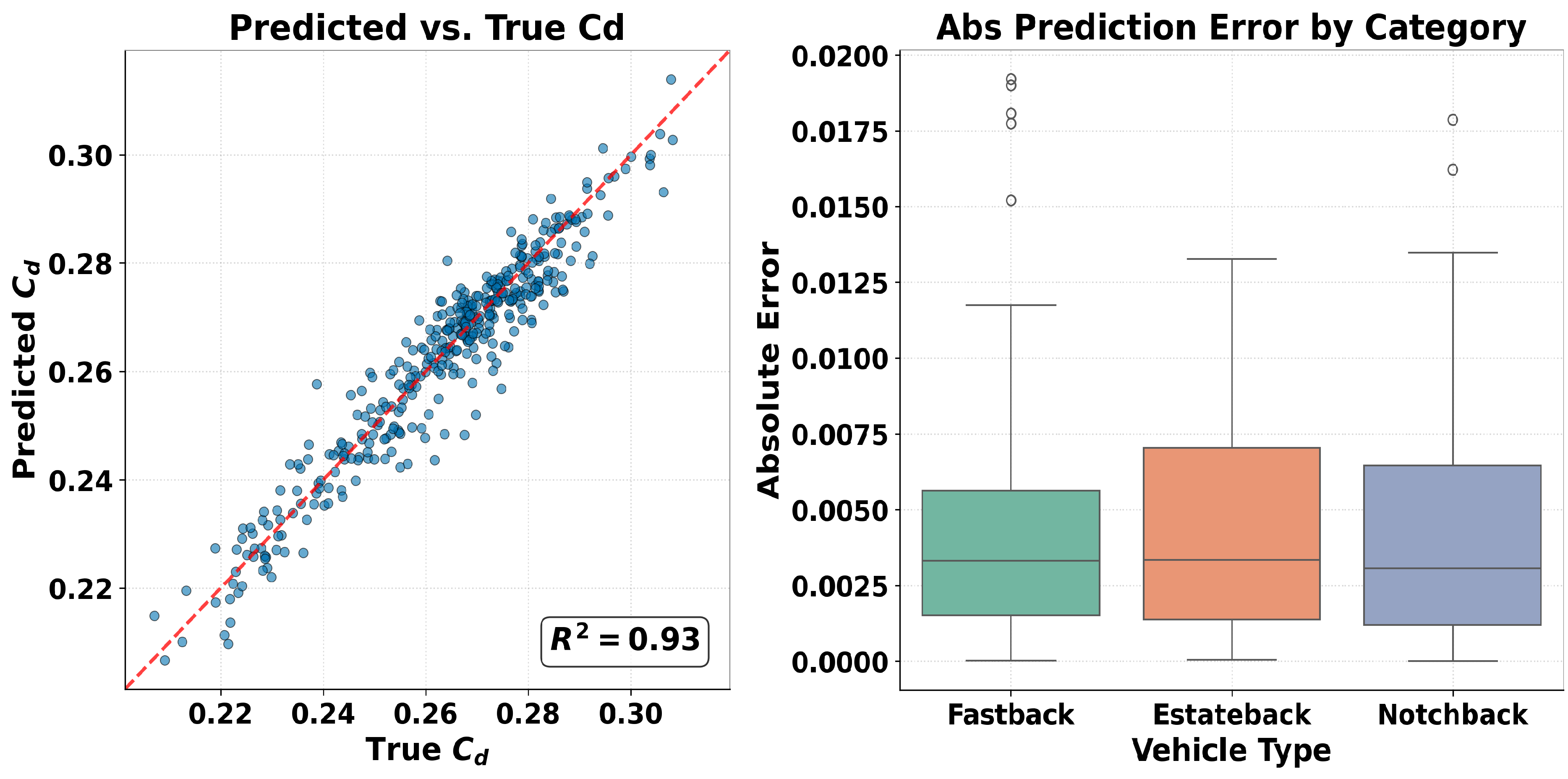}
        \caption{Analysis of $C_d$ prediction performance for the primary model configuration.}
        \label{fig:cd_results_visual_revised}
    \end{subfigure}%
    \hfill
    \begin{subfigure}[t]{0.333\linewidth}
        \centering
        \includegraphics[width=\linewidth]{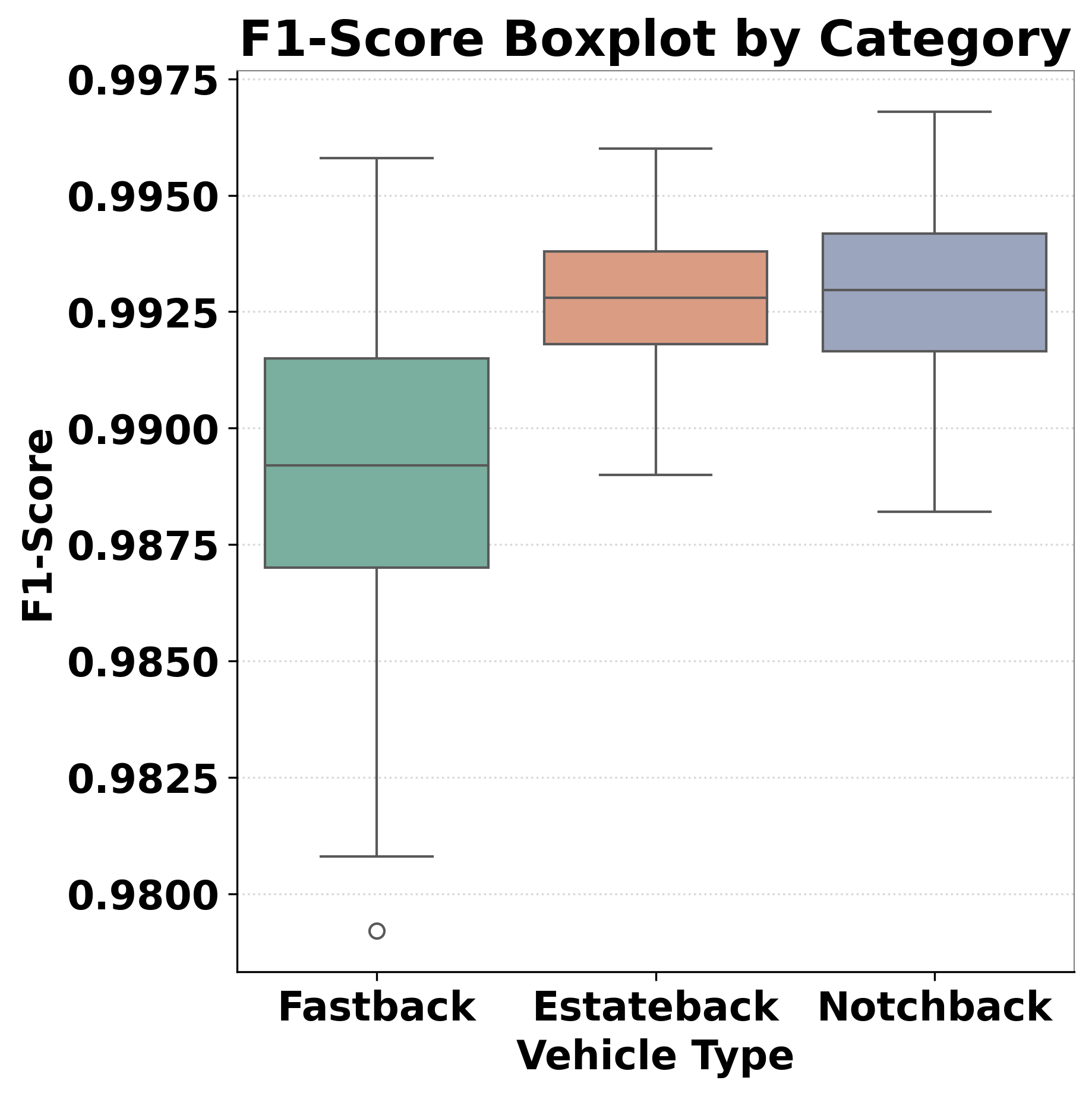}
        \caption{3D shape reconstruction quality for the main model (semi-occupancy).}
        \label{fig:shape_reconstruction_examples_revised}
    \end{subfigure}
    \caption{Performance comparison of the main model across aerodynamic prediction and geometry reconstruction tasks. Left: $C_d$ prediction accuracy. Right: shape reconstruction output.}
    \label{fig:performance_combined}
\end{figure*}

The primary model, trained to simultaneously predict $C_d$ and reconstruct geometry, demonstrated high accuracy. The model achieved a strong coefficient of determination ($R^2$) of 0.93, indicating it explains 93\% of the variance in the true $C_d$ values. The prediction accuracy is further confirmed by low error metrics: a Mean Absolute Error (MAE) of 0.004, a Root Mean Squared Error (RMSE) of 0.005, and a Mean Absolute Percentage Error (MAPE) of only 1.5\%. These values, detailed in Table \ref{tab:cd_prediction_results_main_revised}, confirm that the model's predictions are very close to the CFD-derived ground truth. The definitions for all evaluation metrics are provided in Appendix \ref{app:metrics}. A scatter plot of predicted versus true $C_d$ values (Figure \ref{fig:cd_results_visual_revised}) visually confirms this strong correlation, with most points clustering tightly around the ideal $y=x$ line.

The prediction performance is further explored by examining performance across different vehicle topologies (Figure \ref{fig:cd_results_visual_revised}). The error values are comparable across estate ('E'), fastback ('F'), and notchback ('N') types.

\begin{table}[!htbp]
\centering
\small
\caption{$C_d$ Prediction Performance (Test subset - Simultaneous Training, Semi-Continuous Occupancy).}
\label{tab:cd_prediction_results_main_revised}
\begin{tabular}{@{}lr@{}}
\toprule
\textbf{Metric} & \textbf{Value} \\ \midrule
Number of Samples & $\sim$400, \SI{5}{\percent} Total samples \\
$R^2$ Score & 0.930 \\
Mean Absolute Error & 0.004 \\
Root Mean Squared Error & 0.005 \\
Mean Absolute Percentage Error & $\sim$\SI{1.5}{\percent} \\
Maximum Absolute Error & 0.019 \\ \bottomrule
\end{tabular}
\end{table}



\subsection{Geometric Reconstruction Fidelity}

The model's capacity for accurate 3D shape reconstruction is integral to its aerodynamic prediction ability, as the learned geometric features directly inform the $C_d$ estimation. Furthermore, it enables the validation of the final optimized output using CFD simulations. The high F1 score for classifying interior points (Table \ref{tab:training_ablation_results_main_revised}, main model) indicates that the triplane-based decoder effectively reproduces the vehicle's volume which implicitly defines the surface boundary. Box plots of F1 scores grouped by vehicle type (Figure~\ref{fig:shape_reconstruction_examples_revised}) reveal whether reconstruction quality varies systematically with vehicle typology, similar to $C_d$ prediction. Ideally, high reconstruction fidelity should be maintained across all categories, which is the case here.

\begin{figure}[!htbp]
    \centering
    \includegraphics[width=\linewidth]{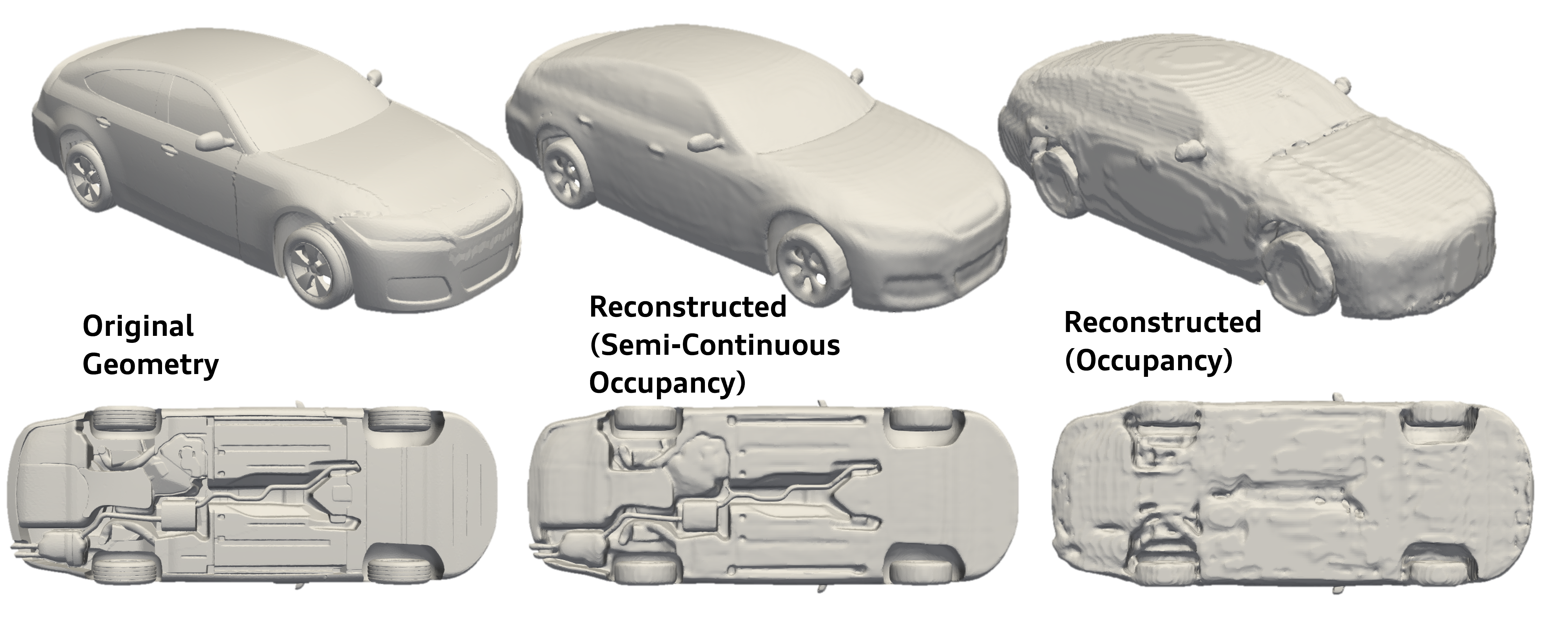}
    \caption{Reconstruction quality for a selected Fastback vehicle with a detailed underbody from the DrivAerNet++ dataset. Original (left), semi-continuous occupancy (middle), and binary occupancy (right) reconstructions.}
    \label{fig:shape_reconstruction_examples_revised2}
\end{figure}

\subsection{Influence of Training Strategies and Occupancy Representation}

In order to validate present work's model design choices and analyze their impact on performance, a series of ablation studies were conducted. These experiments focused on two key aspects: the training regimen and the formulation of the geometric supervision signal.
Two distinct training regimens were evaluated: the primary simultaneous (end-to-end) approach and a sequential two-phase strategy. In the sequential method, the autoencoder components were first trained for geometric reconstruction. Subsequently, the autoencoder weights were frozen, and only the $C_d$ prediction head was trained. The results, presented in Table \ref{tab:training_ablation_results_main_revised}, indicate that the sequential strategy led to significant training instabilities, even though it offered a marginal improvement in $C_d$ prediction metrics. In contrast, the simultaneous approach provided a more stable training process with only a negligible effect on the final reconstruction F1 score. This outcome suggests that joint optimization allows the model to learn a more robust and balanced latent representation that effectively serves both the reconstruction and prediction tasks.

The formulation of the supervision signal for geometric reconstruction proved to be a determining factor in model performance. Current work's proposed semi-continuous occupancy representation was compared against a baseline using absolute binary occupancy values (0 or 1). This semi-continuous target is derived from Signed Distance Functions, as detailed in Equation \ref{eq:sdf_to_occ_revised_main} and Appendix \ref{app:preprocessing}. As shown in Table \ref{tab:training_ablation_results_main_revised}, training with binary targets caused a substantial degradation in performance for both tasks. The $R^2$ value for $C_d$ prediction fell sharply from 0.930 to 0.833. Simultaneously, error metrics increased considerably, with the Mean Absolute Error rising by 94\% from 0.0032 to 0.0062. Geometric fidelity, measured by the F1 score, also deteriorated significantly, dropping by nearly 16\% from 0.992 to 0.836. A qualitative example of the comparison between different supervision signal effect on reconstruction is shown in Figure \ref{fig:shape_reconstruction_examples_revised2}. This performance gap highlights the importance of the smooth gradients provided by the semi-continuous occupancy targets. These gradients are crucial for the implicit triplane decoder to learn an accurate and detailed surface representation. An accurate geometric understanding, in turn, enables the encoder to capture subtle features essential for reliable $C_d$ prediction. The binary targets, lacking this smooth transition near the surface, hinder the network's ability to perform precise surface localization, leading to a coarser geometric representation and a consequently less informative latent space for aerodynamic inference.

\begin{table}[!htbp]
\centering
\caption{Comparative training performance. Simultaneous training provides the best balance of accuracy and stability, outperforming the unstable sequential method and the poorly performing binary occupancy approach.}
\label{tab:training_ablation_results_main_revised}
\small
\resizebox{\linewidth}{!}{%
\begin{tabular}{@{}lrrr@{}}
\toprule
\multirow{2}{*}{\textbf{Metric}} & \textbf{Single-phase training} & \textbf{Two-phase training} & \textbf{Absolute occupancy} \\
 & \textbf{(Simultaneous)} & \textbf{(Sequential)} & \textbf{(Simultaneous, Binary)} \\ \midrule
\multicolumn{4}{l}{\textbf{$C_d$ prediction}} \\
\quad $R^2$ score & 0.9304 & 0.9321 & 0.8328 \\
\quad MAE & 0.0032 & 0.0031 & 0.0062 \\
\quad RMSE & 0.0045 & 0.0043 & 0.0078 \\
\quad MAPE & \SI{1.49}{\percent} & \SI{1.42}{\percent} & \SI{2.38}{\percent} \\
\midrule
\multicolumn{4}{l}{\textbf{Geometric reconstruction}} \\
\quad F1 score inside & 0.9922 & 0.9919 & 0.8363 \\
\midrule
\textbf{Training stability} & Stable & Prone to instability & Stable \\
\bottomrule
\end{tabular}%
}
\end{table}

For geometric reconstruction, the quality is assessed by evaluating the model's ability to correctly classify query points as being inside or outside the vehicle geometry. The predicted occupancy values $o_{pred,i}$ are typically thresholded (e.g., at 0.5) to obtain binary classifications, which are then compared against the binary ground truth occupancy $o_{binary\_true,i}$.
The F1 score is the harmonic mean of precision and recall. It is used here to evaluate the classification of query points, particularly those considered to be "inside" the vehicle (positive class).
Let:
\begin{itemize}
    \item $TP$ (True Positives): Number of query points correctly classified as inside.
    \item $FP$ (False Positives): Number of query points incorrectly classified as inside (actually outside).
    \item $FN$ (False Negatives): Number of query points incorrectly classified as outside (actually inside).
\end{itemize}
Precision measures the accuracy of positive predictions:
\begin{equation}
\text{Precision} = \frac{TP}{TP + FP}
\end{equation}
Recall (or Sensitivity) measures the ability of the model to find all the positive samples:
\begin{equation}
\text{Recall} = \frac{TP}{TP + FN}
\end{equation}
The F1 Score is then calculated as:
\begin{equation}
\text{F1 Score} = 2 \times \frac{\text{Precision} \times \text{Recall}}{\text{Precision} + \text{Recall}} = \frac{2TP}{2TP + FP + FN}
\end{equation}
In the paper (Table \ref{tab:training_ablation_results_main_revised}), "F1 Score Inside" refers to this metric when the positive class is defined as points being inside the vehicle geometry. A higher F1 score indicates better reconstruction fidelity for the vehicle's interior volume and, implicitly, its surface boundary.

\subsection{Model Generalization on Unseen Morphologies}
\label{app:generalization}
To further assess the framework's capabilities beyond the diversity of the training dataset, a test study was conducted on a set of geometries derived from a base model not present in the training or validation splits. The purpose of this study was to evaluate the model's ability to predict the aerodynamic consequences of systematic, localized geometric modifications, a crucial capability for effective shape optimization.

\begin{figure*}[!htbp]
    \centering
    \includegraphics[width=\textwidth]{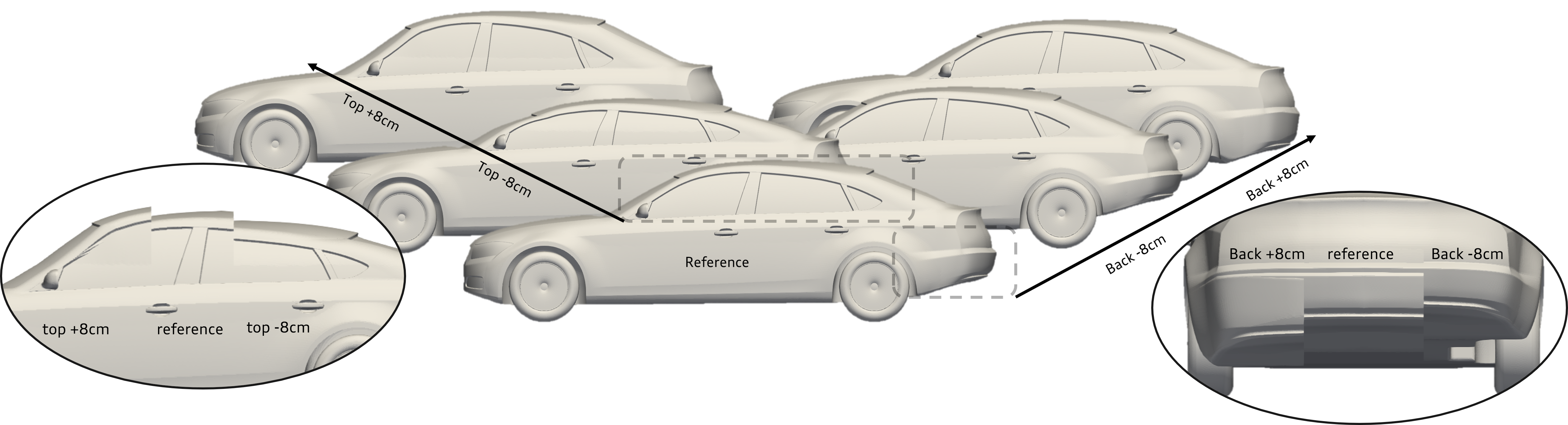}
\caption{Geometries used for the out-of-distribution validation study, showing the reference design (F-S-WWC-WM-001) from the DrivAerNet++ and four systematically morphed variants. Variations include roof height changes (+8\,cm, --8\,cm) and rear-end length changes (+8\,cm, --8\,cm) relative to the reference.}
    \label{fig:morphed_geometries}
\end{figure*}

A baseline fastback geometry, \texttt{F\_S\_WWC\_WM\_001}, was selected. Two distinct regions of the vehicle were targeted for morphing: the roofline and the rear bumper area. Each region was displaced by +8 cm and -8 cm along the local surface normal, creating four new "out-of-distribution" geometries. These four variants, along with the original base geometry, were then analyzed using both our TripOptimizer and independent CFD simulations. Figure \ref{fig:morphed_geometries} provides a visual representation of the base and morphed shapes.

The results, normalized with respect to the drag coefficient of the base geometry, are presented in Table \ref{tab:morphing_validation_results}. The model's predictions demonstrate a strong correlation with the CFD results. Crucially, the model correctly captures the direction of change (the gradient) for every modification. For example, both CFD and the model find that lowering the roof by 8 cm ($T_{n8}$) reduces drag, while raising it ($T_{p8}$) increases drag. Similarly, extending the rear bumper by 8 cm ($B_{p8}$) is correctly identified as beneficial for drag reduction.

The predicted magnitudes of the changes are also in close agreement, with the error between the model predictions and CFD being only 1\% for the most impactful modification ($T_{p8}$) and zero for two of the cases ($B_{p8}$ and $T_{n8}$). 

This strong performance on unseen morphologies indicates that the learned latent space does not merely interpolate between known shapes but captures a more fundamental relationship between geometric features and their aerodynamic effects. The model's ability to accurately sense the $C_d$ gradient with respect to local shape perturbations is a key prerequisite for the optimization strategy detailed in Section \ref{sec:results_discussion_revised}, and these results provide compelling evidence of its validity for guiding design exploration.

\begin{table}[!htbp]
\centering
\small
\caption{Comparison of normalized drag coefficients ($C_d$) for the base geometry and its four morphed variants, as determined by CFD and the data-driven model. Percentage change is relative to the base geometry.}
\label{tab:morphing_validation_results}
\resizebox{\linewidth}{!}{%
\begin{tabular}{@{}lcccc@{}}
\toprule
\multirow{2}{*}{\textbf{Geometry Modification}} & \multicolumn{2}{c}{\textbf{CFD Simulation}} & \multicolumn{2}{c}{\textbf{TripOptimizer Prediction}} \\ \cmidrule(l){2-3} \cmidrule(l){4-5}
 & Normalized $C_d$ & Change (\%) & Normalized $C_d$ & Change (\%) \\ \midrule
Base Reference & 1.00 & ref & 1.00 & ref \\
Rear Bumper -8 cm ($B_{n8}$) & 1.01 & +1\% & 1.02 & +2\% \\
Rear Bumper +8 cm ($B_{p8}$) & 0.98 & -2\% & 0.98 & -2\% \\
Roof -8 cm ($T_{n8}$) & 0.97 & -3\% & 0.97 & -3\% \\
Roof +8 cm ($T_{p8}$) & 1.05 & +5\% & 1.04 & +4\% \\ \bottomrule
\end{tabular}%
}
\end{table}

\subsection{Latent Space Structure and Geometric Similarity}

The model's learned representation of shape variation was evaluated through an analysis of the VAE's latent space. The dimensionality of the latent vectors was reduced for visualization using t-Distributed Stochastic Neighbor Embedding (t-SNE) \cite{hinton2002stochastic}. A perplexity setting of 80 was used. The resulting projections reveal how the model organizes shapes based on both topology and performance-related geometric features.

\begin{figure}[!htb]
    \centering
    \includegraphics[width=\columnwidth]{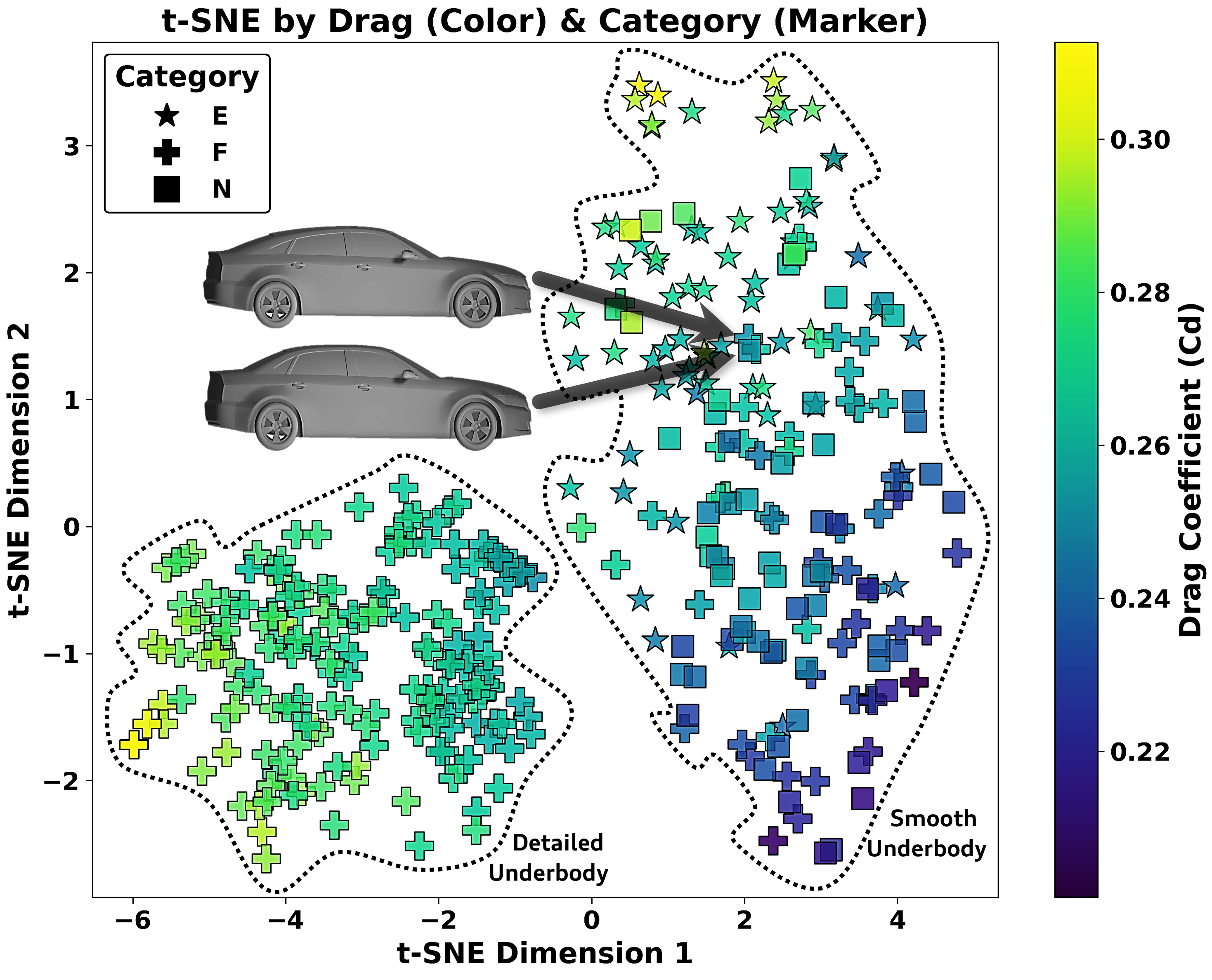}
    \caption{Latent space analysis using t-SNE. Demonstrating geometric similarity grouping and the presence of drag coefficient correlation.}
    \label{fig:tsne_latent_space_analysis}
\end{figure}

Figure \ref{fig:tsne_latent_space_analysis} illustrates the 2D t-SNE projection of the learned latent space. In this visualization, each point represents a vehicle design, with its marker shape corresponding to a predefined vehicle category (e.g., E, F, N) and its color indicating the simulated drag coefficient ($C_d$). The spatial arrangement reveals that the model groups vehicles based on geometric similarity rather than their categorical labels, which exhibit significant overlap. For instance, visually similar car models are mapped to proximate points, demonstrating that the learned representation captures fine-grained shape features. Crucially, this geometric organization is strongly correlated with aerodynamic performance. A clear performance gradient is visible across the latent space, progressing from low-$C_d$ vehicles (dark blue) concentrated in the bottom-right quadrant to high-$C_d$ vehicles (yellow) in the upper and left regions. This confirms that the latent space provides a continuous and physically meaningful embedding of vehicle shapes.

The emergence of this structured and meaningful latent space is fundamental to the optimization strategy. Navigating within this space corresponds to coherent and physically plausible modifications of the decoded vehicle shape. This enables an efficient exploration for improved designs.


\subsection{Aerodynamic Shape Optimization Case Study}
\label{sec:optimization_case_study}
In order to demonstrate the practical application and effectiveness of the proposed encoder parameter refinement strategy, this work presents two optimization case studies. These studies were performed on distinct vehicle topologies from the DrivAerNet++ dataset: a fastback model with detailed underbody and open wheels (F\_D\_WM\_WW\_0864) representing ICE cars, and an estate model with smooth underbody and closed wheels (E\_S\_WWC\_WM\_005) representing the EV category. The objective for both vehicles was to generate a new geometry exhibiting a drag coefficient approximately 10\% lower than its baseline value. The VAE's initial $C_d$ predictions for both geometries showed close agreement with the ground-truth values, establishing a reliable starting point for the optimization. The optimization process consisted of fine-tuning the final three layers of the VAE's encoder over 100 iterative steps, using a learning rate of $1 \times 10^{-5}$. Throughout this procedure, the input point cloud for the initial vehicle was held constant. The model's $C_d$ prediction head provided the gradient signal used to modify the encoder's parameters, thereby steering the latent representation of the shape toward the desired aerodynamic target.

To externally validate the performance of the generated designs, both the initial and optimized geometries were evaluated using independent CFD simulations with more than 150M mesh cells on High-Performance Computing (HPC) cluster. A known challenge in such validation is that absolute $C_d$ values can differ between CFD solvers due to variations in force calculation and normalization. While fundamental domain properties and boundary conditions were kept consistent, this accounts for the minor discrepancy between the baseline $C_d$ values in present work's dataset and those from the validation CFD. The detailed setup for these validation simulations is provided in Appendix \ref{app:cfd_setup}.

The key outcomes are summarized in Table \ref{tab:optimization_results_summary}. The results confirm that the framework achieved substantial aerodynamic improvements. The optimization yielded a CFD-verified drag reduction of 11.8\% for the fastback model and 9.6\% for the estate model, closely aligning with the initial 10\% target.

\begin{table}[!htbp]
\centering
\footnotesize
\caption{Summary of aerodynamic shape optimization case study results.}
\label{tab:optimization_results_summary}
\begin{tabular}{@{}lcc@{}}
\toprule
\textbf{Metric} & \textbf{Fastback car} & \textbf{Estateback car} \\ \midrule
\multicolumn{3}{@{}l}{\textbf{Model-Driven Optimization}} \\
\quad Baseline geometry $C_d$ & 0.289 & 0.249 \\
\quad AI-Optimized geometry $C_d$ & 0.258 & 0.228 \\
\quad Targeted $C_d$ reduction (\%) & 10.0\% & 10.0\% \\ \midrule
\multicolumn{3}{@{}l}{\textbf{CFD Validation}} \\
\quad Baseline geometry $C_d$ & 0.280 & 0.261 \\
\quad AI-Optimized geometry $C_d$ & 0.249 & 0.236 \\
\quad Actual $C_d$ reduction (\%) & 11.8\% & 9.6\% \\
\bottomrule
\end{tabular}
\end{table}

\begin{figure}[h!]
    \centering
    \includegraphics[width=\linewidth]{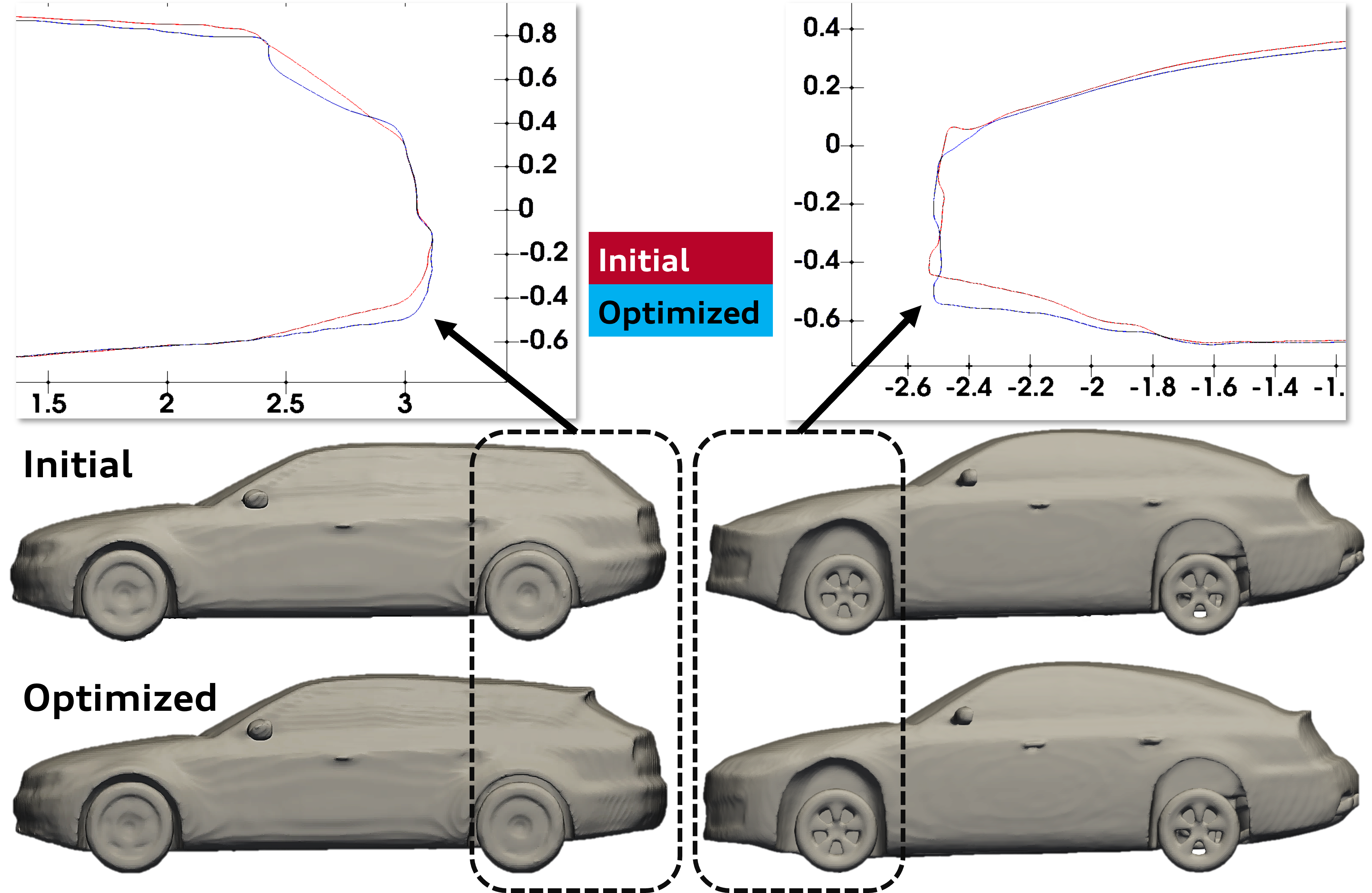}
    \caption{Visual comparison of initial and optimized geometries for the two case studies: a selected Estateback vehicle (left) and a selected Fastback vehicle (right). Top: Overlapped view (Initial - red, Optimized - blue). Middle: Initial geometry. Bottom: Optimized geometry.}
\label{fig:optimization_geometry_comparison}
\end{figure}

\begin{figure}
    \centering
    \includegraphics[width=\linewidth]{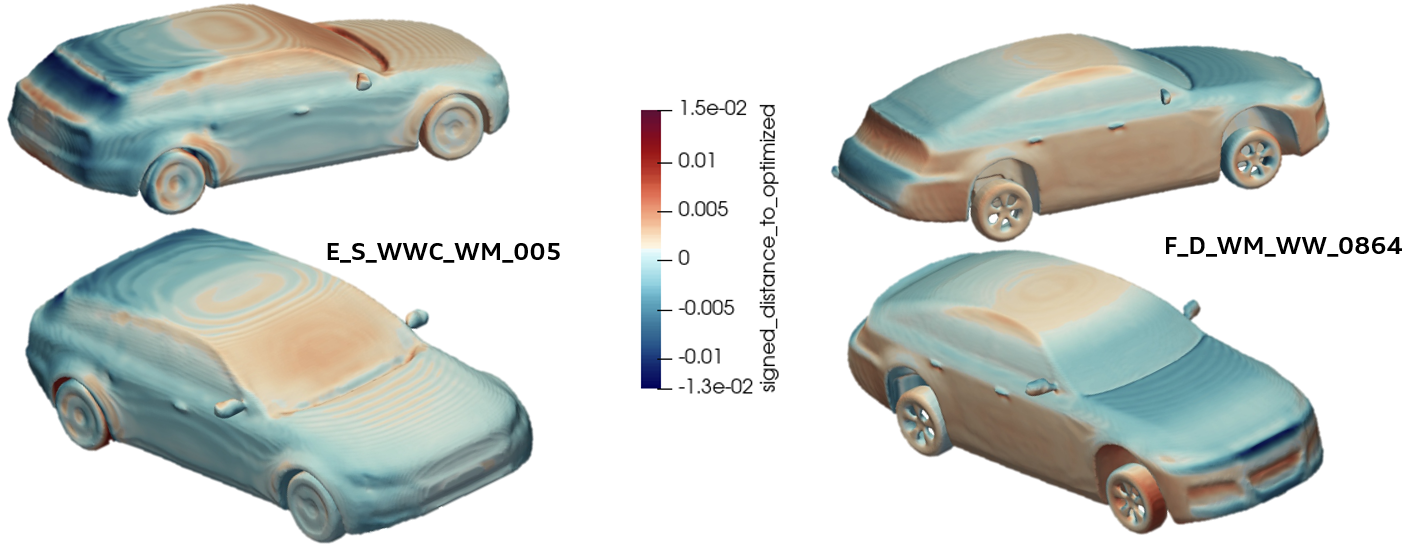}
    \caption{Signed distance maps illustrating the geometric modifications from the initial to the optimized shape for a selected Estateback vehicle (left) and a selected Fastback vehicle (right). The map is overlaid on the initial geometry.}
    \label{fig:optimization_signed_distance}
\end{figure}

The geometric modifications resulting from the optimization are visualized in Figure \ref{fig:optimization_geometry_comparison}. An overlapped view of the initial and optimized shapes reveals subtle yet aerodynamically significant alterations, concentrated primarily in the frontal and rear sections. For a more quantitative analysis, Figure \ref{fig:optimization_signed_distance} presents signed distance maps that visualize the normal displacement between the two surfaces. Positive (red) values indicate outward movement, while negative (blue) values denote inward movement.

These visualizations provide direct, actionable insights into the specific regions modified by the optimization. This level of detail is invaluable for engineering workflows, as it allows designers to translate the data-driven suggestions into high-fidelity CAD models while maintaining full design control and ensuring surface quality. These case studies successfully demonstrate that the encoder refinement technique can effectively guide shape modifications towards improved aerodynamic performance, producing results that are verifiable through standard industry simulation tools.

\textbf{Visualization of Wake Structures Using Total Pressure Coefficient Isosurfaces}: To analyze and compare the wake structures of the initial and optimized car geometries, isosurfaces of the total pressure coefficient \( C_{p,t} \) are visualized, a dimensionless metric representing local total pressure loss relative to the freestream. It is defined as:

\begin{equation}
C_{p,t} = \frac{p_t - p_\infty}{\frac{1}{2} \rho U_\infty^2},
\end{equation}

where \( p_t \) denotes the local total (stagnation) pressure, \( p_\infty \) is the freestream static pressure, \( \rho \) is the freestream density, and \( U_\infty \) is the freestream velocity magnitude. The isosurface corresponding to \( C_{p,t} = 0 \) is used to identify the boundary of regions where the total pressure has dropped to the freestream level, typically indicating areas of significant energy dissipation, flow separation, and wake formation.

\begin{figure*}
    \centering
    \includegraphics[width=\textwidth]{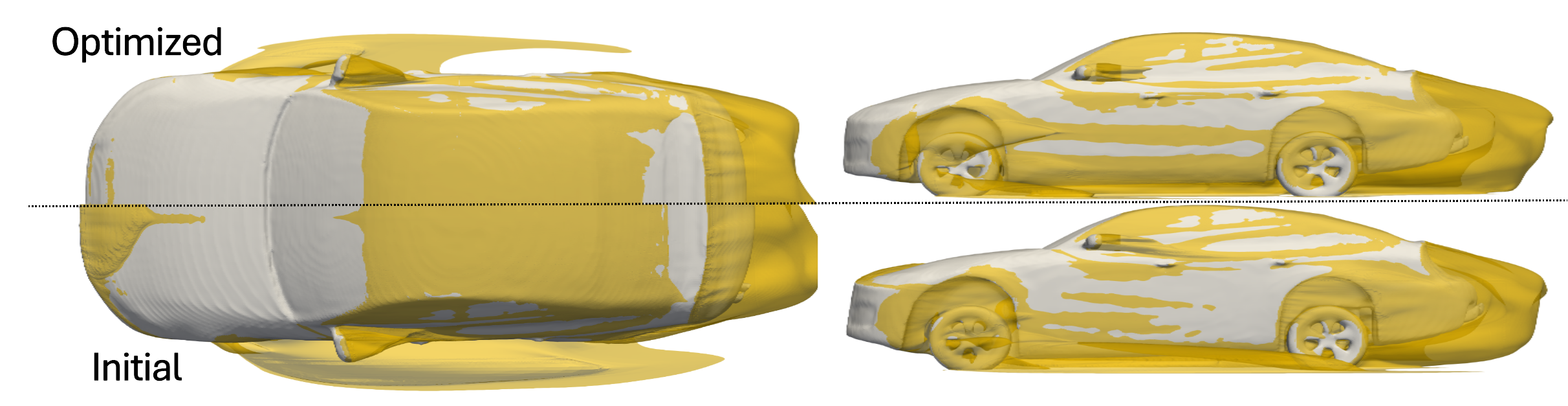}
    \vspace{1em} 
    \includegraphics[width=\textwidth]{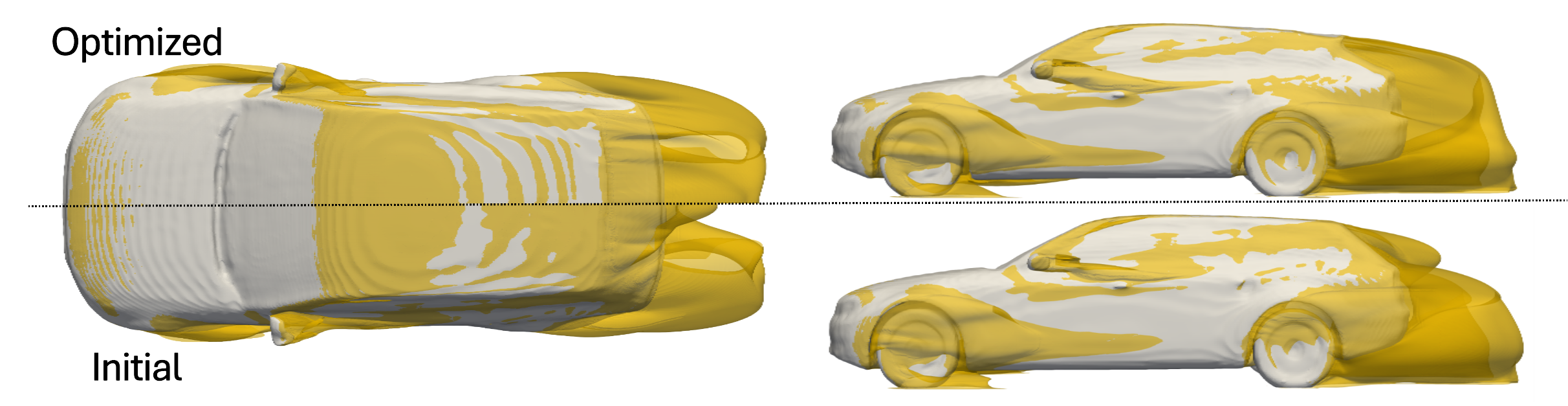}
\caption{Total pressure isosurfaces ($C_{p,t} = 0$) comparing the initial and optimized designs for a Fastback (top) and an Estateback (bottom). Using TripOptimizer, aerodynamic drag was reduced by 11.8\% for the Fastback model and 9.6\% for the Estateback model.}
    \label{fig:optimized_configs}
\end{figure*}

Figure~\ref{fig:optimized_configs} compares the baseline and optimized geometries for two body styles using \( C_{p,t} = 0 \) isosurfaces. The following observations highlight how TripOptimizer adapts specific regions of the vehicle to improve aerodynamic performance:

\begin{itemize}
    \item \textbf{Fastback:} The optimized design shows a visibly reduced wake behind the side mirror and a more streamlined hood profile. Unlike the baseline, the optimized geometry eliminates the flow separation observed on the hood, resulting in improved flow attachment and reduced total pressure loss in the front region.
    
    \item \textbf{Estateback:} The optimization primarily targets the rear end of the vehicle. As a result, the wake region behind the car is more structured and less chaotic compared to the baseline, indicating reduced pressure drag and better control of the rear wake.
\end{itemize}

\textbf{Front-View Pressure Distribution Analysis Using Streamwise Pressure Coefficient}: To assess the frontal pressure distribution and its contribution to form drag, the streamwise pressure coefficient \( C_{p,x} \) was analyzed, defined as:

\begin{equation}
C_{p,x} = \frac{p - p_\infty}{\frac{1}{2} \rho U_\infty^2},
\end{equation}

where \( p \) is the local pressure acting in the streamwise (x) direction, \( p_\infty \) is the freestream static pressure, \( \rho \) is the fluid density, and \( U_\infty \) is the freestream velocity. This dimensionless quantity highlights regions where the flow stagnates or separates, particularly at the vehicle front, and is thus closely associated with pressure drag.

\begin{figure}[!htbp]
    \centering
    \includegraphics[width=\linewidth]{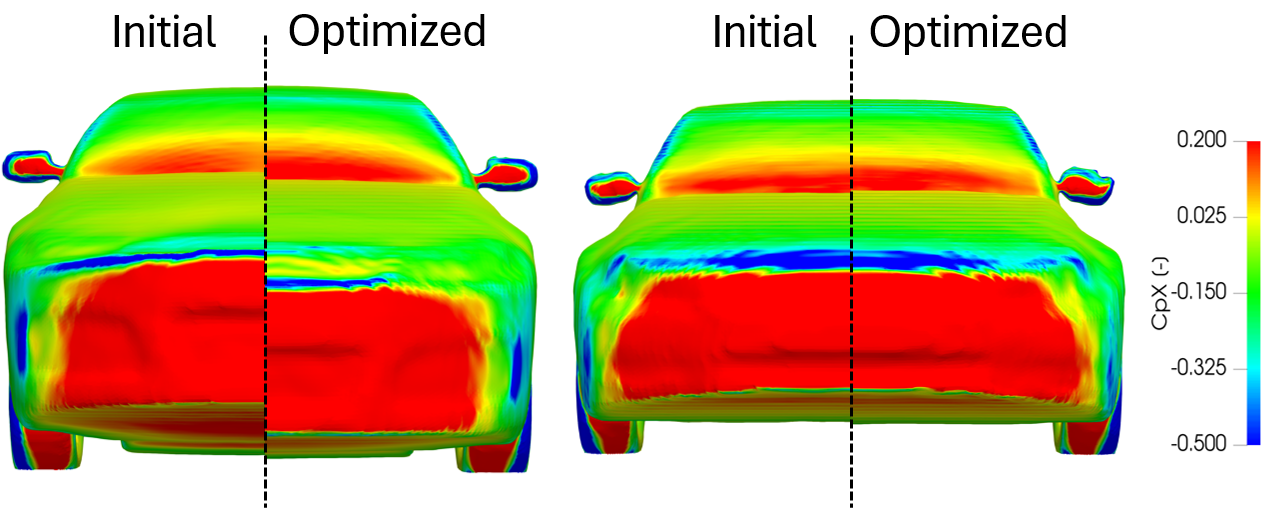}
    \caption{Distribution of \( C_{p,x} \) for the Fastback (left pair) and Estateback (right pair) designs.}
    \label{fig:cpx_optimized}
\end{figure}

Figure~\ref{fig:cpx_optimized} shows the distribution of \( C_{p,x} \) for the Fastback (left pair) and Estateback (right pair) designs, comparing initial and optimized geometries:

\begin{itemize}
    \item \textbf{Fastback:} The optimized design (right half of the split image) exhibits a more symmetric and evenly distributed \( C_{p,x} \) field across the front face, especially around the hood and bumper sides. The reduction in pressure concentration at the upper hood region suggests a more streamlined transition, resulting in improved flow attachment and reduced separation compared to the initial geometry.

    \item \textbf{Estateback:} The optimized design (right) shows a notable reduction in low-pressure (blue) regions near the lower corners of the front bumper. These changes indicate smoother pressure recovery and reduced flow separation in the front-end geometry, which contributes to lowering the total form drag.
\end{itemize}

The collective results from prediction, reconstruction, and optimization case studies successfully validate the effectiveness of TripOptimizer. These findings firmly establish a novel strategy for data-driven aerodynamic shape optimization, capable of generating high-performance designs, validatable by external high-fidelity simulations.



\section{Conclusion}
\label{sec:conclusion}
This paper introduced and validated, TripOptimizer, a fully differentiable deep learning framework based on a Variational Autoencoder for the rapid analysis and optimization of vehicle aerodynamic performance directly from point cloud data. The framework's core is a model combining a triplane-based implicit decoder for high-fidelity 3D shape reconstruction with a dedicated head for precise $C_d$ prediction.

The research demonstrated that the proposed model achieves strong predictive accuracy for $C_d$ ($R^2$ of 0.93) and detailed geometric reconstruction. A key finding highlighted the importance of utilizing semi-continuous occupancy targets derived from signed distance functions for training the shape decoder; this proved crucial for capturing fine surface details essential for accurate aerodynamic analysis. The VAE's learned latent space was shown to effectively organize vehicle shapes based on topologies and more subtle geometric features pertinent to aerodynamics. This work's primary contribution is a novel shape optimization methodology that fine-tunes parameters within the VAE's encoder. This technique allows for targeted geometric modifications to an initial design to achieve a desired $C_d$ value.  The efficacy of this optimization approach was demonstrated through detailed case studies on multiple vehicle topologies, achieving significant $C_d$ reductions of 9.6\% and 11.8\% for the optimized geometries. These results were subsequently validated by independent high-fidelity CFD simulations.

The presented framework provides a valuable and efficient alternative to traditional ASO workflows. It enables a more agile design exploration process by exploring the low-dimensional latent space. The optimization method not only produces improved geometries but can also generate outputs like signed distance maps, which offer actionable insights for engineers to refine original CAD models while retaining design intent and surface quality. Crucially, because the framework operates on point clouds and learns a continuous implicit representation, it is inherently robust to geometric imperfections such as non-watertight surfaces. This represents a practical advantage over traditional adjoint-based ASO, which often fails or requires extensive mesh repair when faced with such common topological issues.

\section{Future Work}
\label{sec:future_work}
The promising results presented in this paper lay the groundwork for several key research directions. A primary objective will be to extend the validation of the TripOptimizer framework across a more diverse range of vehicle topologies beyond the passenger cars in the DrivAerNet++ dataset. This will not only test the model's out-of-distribution generalization but also address its current limitation. A parallel investigation will involve the economic and workflow viability of fine-tuning the TripOptimizer model for new vehicle classes, a key assumption for its projected long-term utility.

Further investigations will also systematically analyze the optimization process itself. This includes studying the sensitivity of the optimization outcome to hyperparameters, such as the number of unfrozen encoder layers, to better understand the trade-off between the magnitude of aerodynamic improvement and the preservation of the initial design's identity. Such analysis will help establish formal constraints to ensure geometric realism during large targeted modifications. To benchmark the proposed encoder-refinement strategy, comparative analyses against both traditional adjoint-based ASO methods and more common latent space traversal techniques are planned.

Finally, a deeper focus will be placed on enhancing the interpretability of the learned representations and the physical mechanisms underlying the model-proposed geometric modifications. This will deepen the engineering insight that can be drawn from the data-driven results. Ultimately, the goal is to integrate TripOptimizer into a multi-agentic framework for holistic vehicle design, where complementary agents handle styling, CAD modeling, meshing, and aerodynamic evaluation, enabling a truly collaborative and accelerated design cycle \cite{elrefaie2025ai}.
\section*{Acknowledgments}
The authors wish to express their sincere gratitude to AUDI AG, particularly the Virtual Development Department, for their invaluable support and for providing the resources and industrial context that were essential for this research. This also includes access to Audi's high-performance computing cluster, which were essential for the training and validation of the deep learning models. We also extend our thanks to the Massachusetts Institute of Technology (MIT), including the Department of Mechanical Engineering and the Schwarzman College of Computing, for fostering to conduct this cross-institutional collaboration. 
\section*{Disclaimer}
The results, opinions, and conclusions expressed in this publication are not necessarily those of Audi Aktiengesellschaft.
\bibliography{references}

\clearpage
\appendix

\section{Appendix}
\subsection{Performance Evaluation Metrics Calculation} \label{app:metrics}

\subsubsection{Aerodynamic Drag Prediction Metrics}
Let $C_{d,true_i}$ be the true drag coefficient for the $i$-th sample,
$C_{d,pred_i}$ be the predicted drag coefficient for the $i$-th sample,
and $N_s$ be the total number of samples in the evaluation set.
The mean of the true drag coefficients is denoted by $\bar{C}_{d,true}$ and is calculated as follows:
\begin{equation}
\bar{C}_{d,true} = \frac{1}{N_s} \sum_{i=1}^{N_s} C_{d,true_i}
\end{equation}

\paragraph{Mean Absolute Error (MAE)}
The MAE measures the average magnitude of the errors in a set of predictions, without considering their direction. It is the average over the test sample of the absolute differences between prediction and actual observation where all individual differences have equal weight.
\begin{equation}
\text{MAE} = \frac{1}{N_s} \sum_{i=1}^{N_s} |C_{d,pred_i} - C_{d,true_i}|
\end{equation}

\paragraph{Mean Squared Error (MSE)}
The MSE measures the average of the squares of the errors. As it squares the errors before averaging, it gives higher weight to larger errors.
\begin{equation}
\text{MSE} = \frac{1}{N_s} \sum_{i=1}^{N_s} (C_{d,pred_i} - C_{d,true_i})^2
\end{equation}

\paragraph{Root Mean Squared Error (RMSE)}
The RMSE is the square root of the MSE. It is interpretable in the same units as the target variable and is also sensitive to large errors due to the squaring term.
\begin{equation}
\text{RMSE} = \sqrt{\frac{1}{N_s} \sum_{i=1}^{N_s} (C_{d,pred_i} - C_{d,true_i})^2}
\end{equation}

\paragraph{Mean Absolute Percentage Error (MAPE)}
The MAPE measures the average absolute percentage difference between the predicted and true values. It is useful for understanding the error relative to the magnitude of the true values.
\begin{equation}
\text{MAPE} = \frac{1}{N_s} \sum_{i=1}^{N_s} \left| \frac{C_{d,true_i} - C_{d,pred_i}}{C_{d,true_i}} \right| \times 100\%
\end{equation}
Note: Care must be taken with MAPE if true values can be zero or very close to zero. In this work, $C_d$ values are strictly positive and generally not close to zero.

\paragraph{Coefficient of Determination ($R^2$ Score)}
The $R^2$ score indicates the proportion of the variance in the dependent variable that is predictable from the independent variable(s). It ranges from $-\infty$ to 1, where 1 indicates perfect prediction, 0 indicates that the model performs no better than constantly predicting the mean of the true values, and negative values indicate worse performance.
\begin{equation}
R^2 = 1 - \frac{\sum_{i=1}^{N_s} (C_{d,true_i} - C_{d,pred_i})^2}{\sum_{i=1}^{N_s} (C_{d,true_i} - \bar{C}_{d,true})^2}
\end{equation}

\subsection{Extended Data Preprocessing Details} \label{app:preprocessing}
This section provides a more granular description of the data preprocessing pipeline, expanding on the methodologies outlined in Section \ref{sec:data_preprocessing}. The implementation relies on libraries such as Trimesh \cite{trimesh} for geometric operations, SciPy for spatial queries, and PyTorch for GPU-accelerated computations. The following steps are applied to each raw STL file to generate the training data.

\subsubsection{Initial Mesh Conditioning}
Before the primary data generation, each loaded mesh undergoes two conditioning steps to ensure robustness and consistency.

\begin{figure*}
    \centering
    \includegraphics[width=\linewidth]{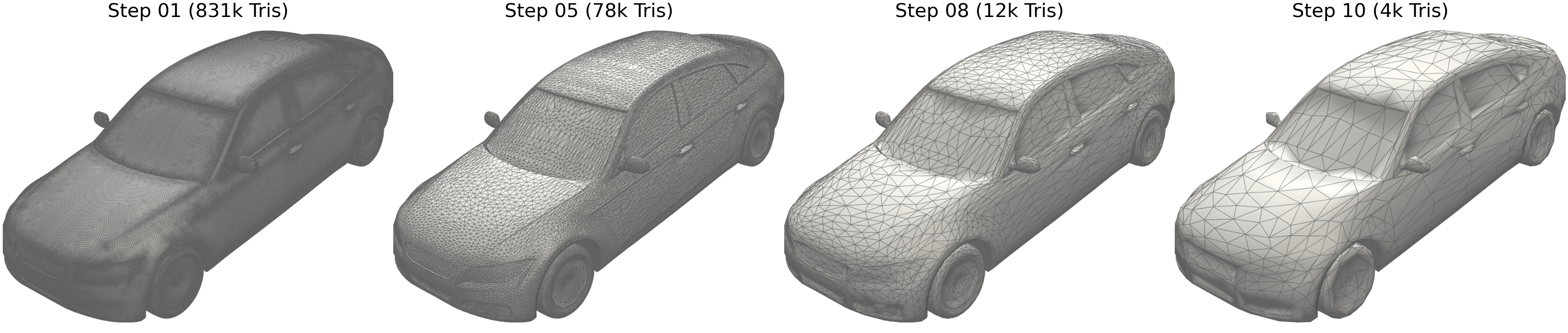}
      \caption{Visual effect of progressive mesh decimation on a DrivAerNet++ vehicle model. (a) Initial high-resolution decimation (Step 01). (b) Moderate decimation (Step 05), maintaining key features. (c) Significant decimation (Step 08), showing increased faceting. (d) Aggressive decimation (Step 10), where feature loss is apparent.}
     \label{fig:decimation_visual_comparison_main_v4}
\end{figure*}

\paragraph{Mesh Decimation}
To manage computational complexity and memory usage for downstream tasks, particularly for meshes with an extremely high polygon count, a decimation step is conditionally applied. If the face count of a mesh exceeds a predefined threshold (e.g., 300,000 faces), it is simplified to a target face count (e.g., 150,000). The primary decimation method utilizes the quadric error metrics algorithm, available through the Open3D` library \cite{Zhou2018}. This algorithm iteratively collapses edges, prioritizing those that introduce the minimal geometric error. The error for collapsing an edge is measured by the sum of squared distances from the new vertex to the planes of its neighboring original faces. This preserves sharp features and overall shape more effectively than uniform simplification methods. If Open3D is unavailable, a fallback vertex clustering decimation is used, where vertices within a specified voxel size are merged. Figure~\ref{fig:decimation_visual_comparison_main_v4} compares mesh appearance across decimation levels. Figures~\ref{fig:decimation_sa_vs_tris_v4} and \ref{fig:decimation_vol_vs_tris_v4} quantify surface area and volume versus triangle count, respectively.

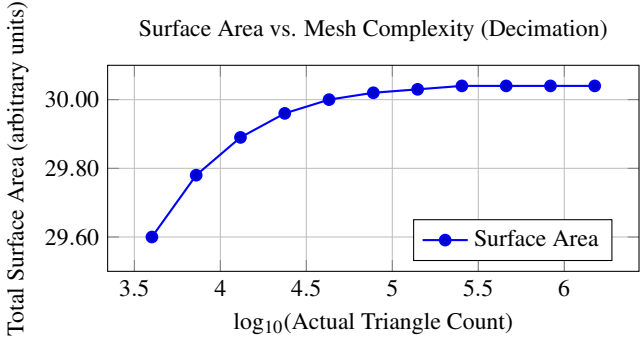
\begin{figure}[!htbp]
    \centering
    \begin{tikzpicture}
        \begin{axis}[
            xlabel={log$_{\text{10}}$(Actual Triangle Count)},
            ylabel={Total Surface Area (arbitrary units)},
            ymin=29.5, ymax=30.1,
            yticklabel style={/pgf/number format/.cd,fixed,fixed zerofill,precision=2},
            legend style={at={(0.95,0.05)},anchor=south east},
            grid=major,
            width=\linewidth,
            height=0.5\linewidth,
            title={Surface Area vs. Mesh Complexity (Decimation)},
        ]
        \addplot+[mark=*, blue, thick] coordinates {
            (6.17712, 30.04) 
            (5.91968, 30.04) 
            (5.66217, 30.04) 
            (5.40465, 30.04) 
            (5.14713, 30.03) 
            (4.88962, 30.02) 
            (4.63212, 30.00) 
            (4.37460, 29.96) 
            (4.11707, 29.89) 
            (3.85950, 29.78) 
            (3.60206, 29.60) 
        };
        \legend{Surface Area}
        \end{axis}
    \end{tikzpicture}
    \caption{Total Surface Area as a function of the logarithm of actual triangle count during progressive mesh decimation. Surface area remains relatively stable initially but decreases more at very low triangle counts.}
    \label{fig:decimation_sa_vs_tris_v4}
\end{figure}
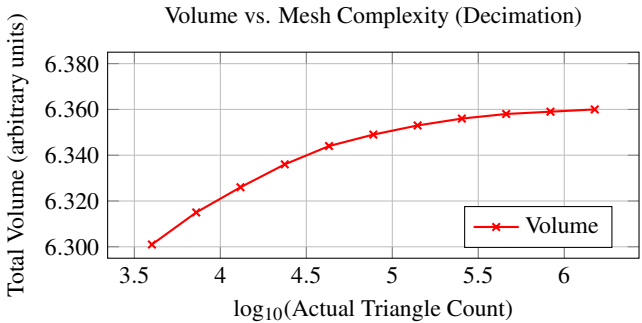
\begin{figure}[!htbp]
    \centering
    \begin{tikzpicture}
        \begin{axis}[
            xlabel={log$_{\text{10}}$(Actual Triangle Count)},
            ylabel={Total Volume (arbitrary units)},
            ymin=6.295, ymax=6.385,
            yticklabel style={/pgf/number format/.cd,fixed,fixed zerofill,precision=3},
            legend style={at={(0.95,0.05)},anchor=south east},
            grid=major,
            width=\linewidth,
            height=0.5\linewidth,
            title={Volume vs. Mesh Complexity (Decimation)},
        ]
        \addplot+[mark=x, red, thick] coordinates {
            (6.17712, 6.360) 
            (5.91968, 6.359) 
            (5.66217, 6.358) 
            (5.40465, 6.356) 
            (5.14713, 6.353) 
            (4.88962, 6.349) 
            (4.63212, 6.344) 
            (4.37460, 6.336) 
            (4.11707, 6.326) 
            (3.85950, 6.315) 
            (3.60206, 6.301) 
        };
        \legend{Volume}
        \end{axis}
    \end{tikzpicture}
    \caption{Total Volume as a function of the logarithm of actual triangle count during progressive mesh decimation. Volume shows a consistent, slight decrease with increasing decimation.}
    \label{fig:decimation_vol_vs_tris_v4}
\end{figure}
\paragraph{Geometric Normalization}
After optional decimation, every mesh is normalized to a canonical coordinate system. This is a critical step to ensure that the neural network receives inputs with a consistent scale and position, which is essential for stable training. The normalization process consists of two parts:
\begin{enumerate}[nosep, leftmargin=*]
    \item \textbf{Centering:} The mesh is translated so that its geometric center aligns with a target coordinate, typically the origin $[0,0,0]$. The geometric center is determined as the center of the mesh's axis-aligned bounding box (AABB). Given a mesh with AABB defined by minimum coordinates $\mathbf{B}_{min}$ and maximum coordinates $\mathbf{B}_{max}$, the translation vector $\mathbf{T}$ is calculated as:
    \begin{equation}
        \mathbf{T} = \mathbf{C}_{target} - \frac{\mathbf{B}_{min} + \mathbf{B}_{max}}{2}
    \end{equation}
    where $\mathbf{C}_{target}$ is the target origin. All vertices $\mathbf{v}_i$ of the mesh are then updated: $\mathbf{v}'_i = \mathbf{v}_i + \mathbf{T}$.

    \item \textbf{Scaling:} The mesh is uniformly scaled to fit within a canonical space. As described in the paper, the method ensures the final bounding box diagonal has a unit length. The scaling factor $S$ is computed based on the extents of the centered mesh, $\mathbf{E} = \mathbf{B}'_{max} - \mathbf{B}'_{min}$:
    \begin{equation}
        S = \frac{d_{target}}{\|\mathbf{E}\|} = \frac{1.0}{\sqrt{E_x^2 + E_y^2 + E_z^2}}
    \end{equation}
    where $d_{target}$ is the target diagonal length (1.0). The final vertex positions are given by $\mathbf{v}''_i = \mathbf{v}'_i \cdot S$.
\end{enumerate}

\subsubsection{Surface Point Cloud Generation}
As mentioned in the paper, the point cloud fed to the VAE encoder is a combination of points sampled based on geometric saliency and points sampled for uniform coverage.

\paragraph{Saliency-Based Sampling}
This method aims to oversample points in regions of high geometric complexity, such as sharp edges and areas of high curvature, which are aerodynamically significant. The process involves several steps:
\begin{enumerate}[nosep, leftmargin=*]
    \item \textbf{Vertex Curvature Estimation:} An approximate curvature value $C_v$ is computed for each vertex $v$. This is derived from the angular deviation between the normals of faces adjacent to the vertex and the average vertex normal itself. The average normal $\bar{\mathbf{n}}_v$ at a vertex is the normalized sum of the normals of its adjacent faces. The curvature is then the average angle:
    \begin{equation}
        C_v = \frac{1}{|\mathcal{F}_v|} \sum_{f \in \mathcal{F}_v} \arccos(\bar{\mathbf{n}}_v \cdot \mathbf{n}_f)
    \end{equation}
    where $\mathcal{F}_v$ is the set of faces adjacent to vertex $v$ and $\mathbf{n}_f$ is the normal of face $f$.

    \item \textbf{Face Saliency Calculation:} A saliency score $S_f$ is computed for each face $f$. This score is a function of the face's area $A_f$ and the curvatures of its vertices, designed to emphasize both large-area faces and high-curvature faces. The formula is:
    \begin{multline}
        S_f = A_f \cdot (w_{edge} \cdot \max_{v \in f}(C_v) \\
        + w_{curv} \cdot \text{mean}_{v \in f}(C_v))^\alpha
    \end{multline}
    where $w_{edge}$ and $w_{curv}$ are weights that balance the influence of sharp edges (max curvature) and curved surfaces (mean curvature), and $\alpha$ is an exponent to control the intensity of the saliency.

    \item \textbf{Probabilistic Sampling:} The set of all face saliency scores is normalized to form a probability distribution $P(f) = S_f / \sum S_f$. A specified number of faces, $N_{saliency}$, are then sampled from the mesh according to this distribution. For each sampled face, a random point is generated on its surface using barycentric coordinates. A point $\mathbf{p}$ on a triangle with vertices $\mathbf{v_0, v_1, v_2}$ is given by:
    \begin{equation}
        \mathbf{p} = w_0\mathbf{v_0} + w_1\mathbf{v_1} + w_2\mathbf{v_2}
    \end{equation}
    where $w_0, w_1, w_2$ are random weights such that $w_i \ge 0$ and $w_0+w_1+w_2=1$.
\end{enumerate}

\begin{figure*}
    \centering
    \includegraphics[width=\linewidth]{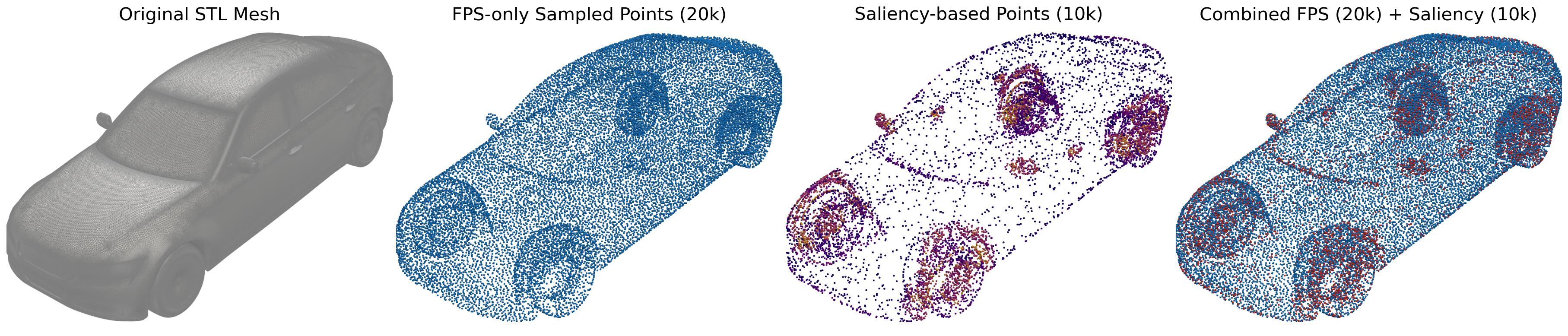}
 \caption{Visual comparison of point cloud sampling strategies applied to a DrivAerNet++ vehicle model. (a) Original geometry. (b) Uniform sampling via FPS. (c) Saliency-based sampling component. (d) Hybrid sampling combining FPS and saliency.}
\label{fig:point_cloud_sampling_comparison_main}
\end{figure*}

\paragraph{Uniform Coverage Sampling via Farthest Point Sampling (FPS)}
To ensure the entire vehicle surface is represented, a set of points is sampled to be maximally distant from each other. This is achieved using the Farthest Point Sampling (FPS) algorithm.
\begin{enumerate}[nosep, leftmargin=*]
    \item \textbf{Initial Surface Sampling:} A dense point cloud of $M$ points (e.g., $M=100,000$) is first sampled uniformly from the mesh surface, weighted by face area.
    \item \textbf{Iterative Selection (FPS):} The FPS algorithm is then applied to this dense set to select a subset of $N_{uniform}$ points. The algorithm proceeds as follows:
    \begin{enumerate}[nosep, leftmargin=*]
        \item Randomly select an initial point $\mathbf{p}_0$ from the dense set and add it to the final sampled set $\mathcal{S}$.
        \item For each point $\mathbf{p}_i$ in the dense set, compute its minimum squared distance $D(\mathbf{p}_i)$ to any point already in $\mathcal{S}$.
        \item Select the point $\mathbf{p}_{farthest}$ from the dense set that has the maximum $D(\mathbf{p}_i)$ value.
        \item Add $\mathbf{p}_{farthest}$ to $\mathcal{S}$ and update the minimum distances $D(\mathbf{p}_i)$ for all points in the dense set with respect to the newly added point.
        \item Repeat steps (c) and (d) until $|\mathcal{S}| = N_{uniform}$.
    \end{enumerate}
\end{enumerate}
The final point cloud for the encoder is the concatenation of the points from the saliency-based and uniform FPS sampling methods. Figure~\ref{fig:point_cloud_sampling_comparison_main} contrasts FPS, saliency-based, and hybrid sampling.

\subsubsection{Semi-Continuous Occupancy Field Generation}
To supervise the geometry reconstruction, a set of query points is sampled in the space surrounding the vehicle, and a semi-continuous occupancy value is computed for each.

\paragraph{Query Point Sampling Strategy}
As outlined in the paper, a mixed sampling strategy is used to generate $M_{total}$ query points to effectively supervise the learning of the surface, interior, and exterior regions.
\begin{enumerate}[nosep, leftmargin=*]
    \item \textbf{Uniform Sampling:} A fraction of points are sampled uniformly within a bounding box slightly larger (e.g., 1.1x) than the normalized mesh's bounding box. This ensures coverage of the far-field space.
    \item \textbf{Surface Sampling:} A fraction of points are sampled directly on the mesh surface, weighted by face area. These points are crucial for learning the exact boundary.
    \item \textbf{Near-Surface Sampling:} The remaining fraction of points is sampled near the surface. This is done by first sampling points on the surface and then perturbing them along their local surface normal direction with Gaussian noise:
    \begin{equation}
        \mathbf{p}_{near} = \mathbf{p}_{surface} + \mathbf{n}_{surface} \cdot \mathcal{N}(0, \sigma^2)
    \end{equation}
    where $\sigma$ is a standard deviation set as a small fraction (e.g., 1.5\%) of the mean extent of the normalized bounding box. This helps the network learn the gradient of the occupancy field near the boundary.
\end{enumerate}

\paragraph{Occupancy Value Calculation}
For each of the $M_{total}$ query points $\mathbf{x}_q$, a Signed Distance Function value is first estimated, which is then mapped to a semi-continuous occupancy value.
\begin{enumerate}
    \item \textbf{Unsigned Distance:} The magnitude of the SDF is approximated by finding the Euclidean distance from the query point $\mathbf{x}_q$ to the closest vertex on the mesh. This is performed efficiently for all query points using a k-d tree constructed from the mesh vertices.

    \item \textbf{Sign Determination:} The sign of the SDF (inside or outside) is determined using a robust majority-vote ray-casting method. For each query point $\mathbf{x}_q$, a small number of rays (e.g., 6) are cast in random directions. For each ray, the number of intersections with the mesh is counted.
    \begin{equation}
    \text{vote}(\mathbf{x}_q, \mathbf{d}_i) =
    \begin{cases}
      -1 & (\text{inside}), \text{ if } |\text{intersections}(\mathbf{x}_q, \mathbf{d}_i)| \text{ is odd} \\
      +1 & (\text{outside}), \text{ if } |\text{intersections}(\mathbf{x}_q, \mathbf{d}_i)| \text{ is even}
    \end{cases}
    \end{equation}
    The final sign for $\mathbf{x}_q$ is determined by the sign of the sum of votes across all ray directions. This method provides robustness against non-watertight meshes or other geometric imperfections.

    \item \textbf{SDF to Occupancy Mapping:} The computed SDF value is transformed into a semi-continuous occupancy value $o_{true}$ using the clipping function from Equation \ref{eq:sdf_to_occ_revised_main}:
    \begin{equation}
        o_{true}(\mathbf{x}_{q}) = \text{clip}_{[0,1]}\left(0.5 - 0.5 \cdot \frac{\text{SDF}(\mathbf{x}_{q})}{s}\right)
    \end{equation}
    Here, the smoothing parameter $s$ is dynamically set as a fraction (e.g., 3\%) of the mean extent of the normalized mesh's bounding box. This creates a smooth transition from 1 (deep inside) to 0 (far outside) in a narrow band around the vehicle surface, providing a richer gradient signal for the neural network compared to a binary occupancy field.
\end{enumerate}

\subsection{Hyperparameter Specifications} \label{app:hyperparams}
This section provides the specific hyperparameter values used for data preprocessing, model architecture, and the training process, as referenced in the main text.

Table \ref{tab:data_preprocessing_hyperparams} outlines the parameters used for preparing the geometric data for model consumption. This includes the specifics of point cloud sampling from the mesh surfaces and the generation of the semi-occupancy field used for training the decoder.

\begin{table}[!htbp]
\centering
\footnotesize
\caption{Key Data Preprocessing Hyperparameters.}
\label{tab:data_preprocessing_hyperparams}
\begin{tabular}{ll}
\toprule
\textbf{Parameter} & \textbf{Value/Description} \\ \midrule
\multicolumn{2}{l}{\textbf{Surface Point Cloud Generation}} \\
\multicolumn{2}{l}{\textbf{for Encoder}} \\
Sampling Strategy & 75\% FPS, 25\% high curvature/edges \\
Number of Surface Points ($N$) & 50,000 \\ \midrule
\multicolumn{2}{l}{\textbf{Semi-Occupancy Field Generation}} \\
Total Query Points per Mesh ($M_{total}$) & 1,280,000 \\
Query Point Sampling Ratios: & \\
\quad Uniform / Surface / Near-Surface & 0.50 / 0.25 / 0.25 \\
Near-Surface Perturbation StdDev & 0.015 (of bounding box extent mean) \\
Occupancy Threshold Parameter ($s$) & 0.03 (of normalized bbox extent mean) \\
Mesh Normalization Target & Unit diagonal length \\
Raycast Directions for Sign & 6 (random) \\ \midrule
\multicolumn{2}{l}{\textbf{Dataloader Subsampling}} \\
\multicolumn{2}{l}{\textbf{for VAE Training}} \\
Domain Query Points for Decoder ($M$) & 131,072 \\ \bottomrule
\end{tabular}
\end{table}
The architectural hyperparameters for the main components of our VAE model are detailed in Table \ref{tab:model_architecture_hyperparams}. These values define the structure and capacity of the point cloud encoder, the triplane occupancy decoder, the occupancy MLP, and the final drag coefficient ($C_d$) prediction head.

\begin{table}[!htbp]
\centering
\caption{Key Model Architecture Hyperparameters.}
\label{tab:model_architecture_hyperparams}
\resizebox{\columnwidth}{!}{%
\begin{tabular}{@{}llp{4cm}@{}}
\toprule
\textbf{Component} & \textbf{Parameter} & \textbf{Value} \\ \midrule
\multirow{3}{*}{\textbf{General Latent Space}} & Latent Dimension ($D_{latent}$) & 128 \\
& Latent Grid Resolution ($H_{latent} \times W_{latent}$) & $24 \times 24$ \\
& Fourier Num Frequencies ($L$) & 9 \\ \midrule
\multirow{3}{*}{\textbf{Point Cloud Encoder}} & ResNet Block Channels & [256, 1024, 1024, 1024] \\
& Final Attention Aggregation Heads & 16 \\
& Final Attention Dropout & 0.1 \\ \midrule
\multirow{4}{*}{\textbf{Triplane Occupancy Decoder}}
& Triplane Feature Dimension ($C_{plane}$) & 128 \\
& U-Net Block Output Channels & [512, 256, 128, 64] \\
& U-Net Layers per UpDecoderBlock & 2 \\
& U-Net Norm Num Groups & 32\\ \midrule
\multirow{3}{*}{\textbf{Occupancy MLP}} & Input Features (from Triplanes) & $3 \times C_{plane} = 384$ \\
& Hidden Dimension & 128 \\
& Num Layers & 4 \\ \midrule
\multirow{5}{*}{\textbf{$C_d$ Prediction Head}} & ConvNet Channels & [256, 128, 64, 32] \\
& ConvNet SE Reduction Factor & 16 \\
& Self-Attention Heads & 8 \\
& MLP Embed Dimension & 256 \\
& MLP Depth & 4 \\
\bottomrule
\end{tabular}%
}
\end{table}

Finally, Table \ref{tab:training_hyperparams} lists the settings used during the model training phase. This covers the optimizer configuration, learning rate schedule, and the specific weights applied to the different components of our composite loss function.

\begin{table}[!htbp]
\centering
\caption{Key Training Hyperparameters.}
\label{tab:training_hyperparams}
\begin{tabular}{@{}ll@{}}
\toprule
\textbf{Parameter} & \textbf{Value} \\ \midrule
Learning Rate & $5 \times 10^{-5}$ \\
Weight Decay & $1 \times 10^{-6}$ \\
Scheduler & StepLR \\
Scheduler Step Size & 5 epochs \\
Scheduler Gamma & 0.75 \\
Number of Epochs & 80 \\
Optimizer & AdamW \\ \midrule
\multicolumn{2}{l}{\textbf{Loss Function Configuration}} \\
Reconstruction Loss Type & Smooth L1 \\
Smooth L1 Beta ($\beta$) & 0.05 \\
KL Start Weight & $1 \times 10^{-9}$ \\
KL Target Weight & $4 \times 10^{-9}$ \\
KL Annealing Epochs & 5 \\
Reconstruction Weight ($w_{recon}$) & 1.0 \\
$C_d$ Prediction Weight ($w_{C_d}$) & 10.0 \\
Boundary Loss Weight ($w_{boundary}$) & 1.0 \\
Triplane L1 Reg. Weight ($w_{triplane\_L1}$) & $5 \times 10^{-4}$ \\
\bottomrule
\end{tabular}
\end{table}
\subsection{CFD Validation Setup for Optimization Case Study} \label{app:cfd_setup}
This section details the setup parameters used for the independent Computational Fluid Dynamics (CFD) simulations performed to validate the optimized vehicle geometry discussed in Section \ref{sec:optimization_case_study}. The simulations were conducted using OpenFOAM \cite{jasak2007openfoam}, an open-source CFD software package. Due to the proprietary nature of the project, detailed images of the mesh and computational domain cannot be shared. The setup was designed to replicate standard automotive aerodynamic testing conditions.

\subsubsection{Solver Details}
The simulations employed a steady-state solver for incompressible, turbulent flows. Specifically, the `simpleFoam`\cite{jasak2007openfoam} solver was used, which is based on the SIMPLE (Semi-Implicit Method for Pressure-Linked Equations) algorithm. Each simulation was executed for up to 4000 iterations to ensure convergence of the solution. The final aerodynamic coefficients were computed by averaging the flow field values over the last 500 iterations. The fluid was defined with a density $\rho$ of 1.1584 kg/m$^3$ and a dynamic viscosity $\mu$ of $1.82 \times 10^{-5}$ Pa$\cdot$s. The simulations were run in parallel on 512 CPU cores, with each case requiring approximately 7 hours of computation time.

\subsubsection{Meshing}
The computational mesh for the virtual wind tunnel was generated using the `snappyHexMesh` utility within OpenFOAM. This utility created a body-fitted mesh of approximately 150 million cells from the vehicle's STL surface representation. The mesh was primarily composed of hexahedral cells with a base cell size of 1.5 meters in the far-field.

To accurately capture the flow physics, extensive local refinement was applied. Multiple refinement volumes were defined around the vehicle, with mesh levels increasing progressively closer to the car body. For instance, the region immediately surrounding the car was refined to level 10, while aerodynamically critical components such as the wheel housings level 11. Prism layers were extruded from the vehicle's no-slip surfaces to properly resolve the boundary layer and ensure the fidelity of the wall shear stress and pressure predictions. The overall mesh quality was high, with the non-orthogonality value remaining below 65 degrees for the vast majority of cells (+99.9\%).

\subsubsection{Turbulence Model}
A Reynolds-Averaged Navier-Stokes (RANS) approach was chosen to model the effects of turbulence. The specific turbulence model employed was the two-equation $k-\omega$ SST (Shear Stress Transport) model\cite{wilcox1988reassessment,menter1994two}. This model is widely used and validated for external vehicle aerodynamics as it combines the robustness and accuracy of the $k-\omega$ model in near-wall regions with the freestream independence of the $k-\epsilon$ model in the far-field, making it well-suited for predicting flow separation and aerodynamic forces.

\subsubsection{Boundary Conditions}
Standard boundary conditions for an automotive external aerodynamics simulation were applied to the computational domain:
\begin{itemize}[nosep, leftmargin=*]
    \item \textbf{Inlet:} A uniform velocity of 30 m/s (108 km/h) was specified at the domain inlet, with a turbulence intensity of 0.5\%.
    \item \textbf{Outlet:} A zero-gradient condition was applied for velocity, and a fixed static pressure of 0 Pa (relative) was set at the outlet.
    \item \textbf{Ground:} A moving wall condition was applied to the ground plane, with its velocity matching the inlet velocity of 30 m/s. This simulates the relative motion between the vehicle and the road, preventing the formation of an unrealistic boundary layer under the car.
    \item \textbf{Vehicle Surfaces:} A no-slip wall condition was applied to all surfaces of the vehicle geometry.
    \item \textbf{Symmetry and Far-Field:} Depending on the setup, symmetry or slip conditions were applied to the top and side boundaries of the virtual wind tunnel to simulate an unconfined airflow.
\end{itemize}

\subsubsection{Computational Domain}
The simulations were performed within a large rectangular computational domain ($72 \times 60 \times 36$ m) representing a virtual wind tunnel to minimize blockage effects and ensure fully developed flow around the vehicle. The vehicle was positioned to provide sufficient distance to the domain boundaries, typically several vehicle lengths upstream and downstream and multiple vehicle widths to the sides and top. The frontal area used for calculating the aerodynamic coefficients was based on the actual frontal value of each geometry.

\end{document}